\documentclass[11pt]{article}
\usepackage[T1]{fontenc}
\usepackage[utf8]{inputenc}
\usepackage{lmodern}
\usepackage[english]{babel}
\usepackage{amsmath,amssymb,amsthm}
\usepackage{booktabs}
\usepackage{graphicx}
\usepackage{microtype}
\usepackage[margin=1.15in]{geometry}
\usepackage{natbib}
\usepackage[hidelinks]{hyperref}
\usepackage{xcolor}
\usepackage{xurl}

\newtheorem{proposition}{Proposition}
\newtheorem{definition}{Definition}
\newtheorem{remark}{Remark}

\title{\textbf{Across-Design Uncertainty in Short Pricing Panels: Inference and Identification}}

\author{Pedro Cadahia Delgado\thanks{Corresponding author.} \\
\small PhD, Universidad de Huelva, Spain}

\date{}

\begin{document}
\maketitle

\begin{abstract}
\noindent
Short observational pricing panels often contain many observations but few distinct price movements. We evaluate the inferential consequences of this sparsity in a synthetic data-generating process by separating estimation error into uncertainty conditional on a realized price trajectory and variation across alternative trajectories. In baseline simulations, this across-design component accounts for $97.6\%$ of estimation error variance for a gradient-boosted specification, causing coverage shortfalls driven by design-specific centering error that standard within-panel resampling and cluster-robust procedures fail to capture.

Three main results organize the analysis. First, across-design dispersion follows the empirical relation $\hat\sigma_b \approx 0.182 V^{-0.271}$, where $V=n_{\text{moves}}\times\text{magnitude}^2$, with $-0.271$ treated as a simulation regularity. Second, adding regions sharing a common price path improves nuisance estimation but creates no independent price trajectories; only averaging across units with independent design errors reduces across-design standard deviation at the $\sqrt{k}$ rate. Third, a Paule--Mandel variance component estimated across independently priced units increases empirical coverage under homogeneity from $0.469$ to $0.931$. Broadly, improving inference in passive panels requires generating independent identifying variation, such as through controlled regional randomization.

Finally, an application to scanner data (Dominick's Finer Foods, Soft Drinks) confirms these findings: nominal price zones and products behave as a small fraction of their count in independent design draws, yielding between-unit dispersion intervals far wider than conventional within-panel bootstraps.


\end{abstract}

\clearpage
\section{Introduction}\label{sec:intro}

Empirical pricing work often reports an elasticity estimate together with a confidence interval constructed from the realised panel. In short panels, however, the number of rows can be a poor description of the amount of treatment variation available for identifying the price coefficient. A manufacturer may observe many region--week cells while changing a common list price only a handful of times. This paper asks what conventional inferential procedures represent in that setting and what additional uncertainty appears when the realised price trajectory itself is treated as one draw from a broader pricing process.

The organizing distinction is between \emph{within-design} and \emph{between-design} uncertainty. Let a design $D$ denote the realised configuration of price movements and other non-shock features of the panel, and let $u$ denote the demand shock. Conditional on $D$, repeated draws of $u$ generate the familiar sampling or shock uncertainty around an estimator. Across alternative realisations of $D$, the estimator can also be displaced because different price trajectories provide different residualised treatment variation and different alignments with confounders. We refer to the dispersion of this design-specific centring error as \emph{between-design uncertainty}. The distinction is conceptually related to the separation between sampling-based and design-based uncertainty in \citet{abadie2020sampling} and to recent design-based analyses of quasi-experiments \citep{rambachan2025design}. That separation itself traces back to the randomization-inference tradition, in which the assignment mechanism---not a hypothesised superpopulation---is the sole source of randomness \citep{neyman1990application, fisher1935design}, and to the subsequent recognition that regression-based estimators computed on one realised assignment can be analysed either conditionally on that assignment or over the distribution of assignments the design could have produced \citep{freedman2008regression, athey2017econometrics}. Our use of the terminology is narrower than this literature in one respect and broader in another: narrower, because we do not study a randomised assignment mechanism, only the DGP's own frequency of generating price trajectories; broader, because the simulations explicitly generate multiple pricing trajectories so that the two components of the estimation error can be measured directly rather than bounded.

This distinction also clarifies what can and cannot be learned from a single realised panel. A block bootstrap, a cluster-robust covariance estimator, or a multiway-clustered covariance estimator can be appropriate for aspects of dependence and repeated-sampling variation supported by the observed data. None of these procedures is designed, by itself, to nonparametrically recover the distribution of estimator displacement across price histories that were not observed. This is not the claim that such methods are generically invalid. It is a statement about the object they condition on and about the additional assumptions or data required to learn a distribution over alternative designs.

We study this issue in a synthetic data-generating process with $120$ weeks, sparse list-price movements, promotions, competitor prices, calendar variables, weather and inflation. The baseline configuration has ten list-price changes per series with magnitudes between $3.0\%$ and $7.5\%$. These values are chosen to be broadly comparable with empirical moments reported by \citet{nakamura2008five}; the calibration is intended to place the exercise in a plausible pricing regime rather than to establish external validity for any particular category. Because the generator supplies the ground truth and permits repeated shocks within the same price trajectory as well as repeated price trajectories, it makes observable a decomposition that cannot be learned nonparametrically from one realised historical panel.

The baseline simulation is deliberately stark. For the gradient-boosted specification, $97.6\%$ of the variance of the estimation error is associated with variation in the design-specific conditional mean error; for the sieve specification the corresponding share is $99.3\%$. The reported bootstrap standard error is larger than the measured within-design standard deviation but much smaller than the across-design dispersion. Thus the empirical finding is not simply that one bootstrap happens to be too narrow. Rather, in these simulations a substantial part of the coverage shortfall is associated with variation in the conditional centring of the estimator across realised price histories.

\paragraph{Contributions.} The paper makes three related contributions.

\begin{enumerate}
\item \textbf{Within-design versus between-design uncertainty.} We formulate the variance decomposition on the estimation error $\hat\theta-\theta(D)$ and measure its two components by holding the design fixed while resampling shocks and then repeating the exercise across designs. In the baseline DGP, between-design centring dispersion is much larger than within-design shock dispersion. The exercise shows why procedures based only on one realised panel need not represent repeated-design uncertainty, while avoiding the stronger claim that all within-panel inference is invalid for conditional questions.

\item \textbf{More observations are not the same as more identifying trajectories.} Adding region--week rows under a common national list price may improve estimation of outcome-side objects and reduce some conditional noise, but it does not multiply the number of independently realised price paths. More generally, the variance of an average of design-specific errors depends on their covariance. Under approximately independent, mean-zero design-specific errors, averaging $k$ independently priced units reduces their standard deviation by $\sqrt{k}$; under a common design component, the reduction is smaller and can disappear. This provides the formal basis for the paper's practical message that ``adding rows'' is not equivalent to ``adding identification.''

\item \textbf{From inferential correction to data design.} Across the simulated grid, the relation $\hat\sigma_b \approx 0.182V^{-0.271}$ summarizes how design dispersion changes with the heuristic variation index $V=n_{\text{moves}}\times\text{magnitude}^2$. The exponent is reported only as an empirical regularity of these simulations. A variance-component construction across independently priced units improves repeated-design coverage, but it becomes wide and, when true effects differ, mixes design dispersion with genuine heterogeneity. These findings shift the practical emphasis from searching only for a more elaborate post-hoc interval toward generating additional independent treatment variation. Randomized regional pricing is one transparent design option, not a logically necessary one.
\end{enumerate}

\paragraph{A real-data illustration, and what having both perspectives adds.} None of the three contributions above requires real transactions: the decomposition, the aggregation result, and the frontier and variance-component exercises are all measured on one synthetic generator, which is what makes $\theta(D)$ and repeated designs observable in the first place. Section~\ref{sec:realdata} complements them with an illustrative application of the same pipeline to a real scanner-data category. A generator alone cannot show that the mechanism it isolates is present outside the world built to produce it; a real panel alone cannot supply the repeated, ground-truth designs that the decomposition of Section~\ref{sec:decomposition} requires, since only one price history is ever realised for a given category. Holding both together is therefore not redundant: the simulation identifies and measures a mechanism under conditions engineered to make it observable, and the real category asks whether that mechanism's qualitative signature---correlated nominal units, and a wide gap between an interval that represents between-unit dispersion and one that does not---appears where no ground truth is available to isolate it directly. The two exercises answer different questions, and neither substitutes for the other; \S\ref{sec:reallimits} states explicitly what the real-data illustration cannot establish.

\paragraph{Conditional and repeated-design questions.} The paper reports both conditional coverage given a realised design and coverage averaged over generated designs. These answer different questions. Conditional coverage asks how an interval behaves if the observed price trajectory is held fixed and only the shock is repeated. Repeated-design coverage asks how the full estimation procedure behaves when the price trajectory is itself redrawn from the DGP. The second object is useful for methodological evaluation and for a firm contemplating future pricing histories; it should not be conflated with the first.

\paragraph{Random versus fixed designs.} Whether the price trajectory $D$ is treated as fixed or as a draw from some distribution is a modelling choice with a long history under the heading of random-design versus fixed-design regression \citep{imbensrubin2015causal}, and it is a different question from whether prices were experimentally randomised. A firm that never randomises still generates a new $D$ every time it repeats a pricing episode in a new market, season, or product line; the between-design uncertainty studied here is about that repetition, not about an experimental assignment mechanism. Treating $D$ as fixed---the implicit stance of within-panel resampling and cluster-robust inference---is the right choice for a conditional question about the one trajectory that was observed, and the wrong choice for an unconditional question about how the same pricing process would perform on the next trajectory it generates \citep{athey2017econometrics}. The simulations let both stances be evaluated on the same DGP because $D$ is drawn repeatedly by construction, which is generally not possible with a single observed historical panel.

\paragraph{Relation to weak identification.} Sparse price movement reduces the residualised treatment variation available in the partially linear regression. This is naturally related to the broader weak-identification and weak-signal literature \citep{staigerstock1997instrumental, stock2005testing, andrews2019weak}, but the connection is analogical rather than an application of weak-IV asymptotics. The model here does not introduce a weak instrument. The relevant empirical symptom is that the denominator based on residualised price variation can be small or highly design-dependent, making finite-sample centring and dispersion sensitive to the realised trajectory. We therefore use ``limited identifying variation'' as the primary description and reserve ``weak identification'' for the broader conceptual connection.

\paragraph{Effective variation and effective sample size.} A related literature formalises the idea that the number of rows overstates the information available for a treatment-effect estimate when treatment variation is unevenly spread or overlap is limited, proposing measures of an \emph{effective sample size} that can be far smaller than the nominal one \citep{crump2009dealing, aronowsamii2016does}. The variation index $V=n_{\text{moves}}\times\text{magnitude}^2$ and the residualised-variance share $\operatorname{Var}(\tilde T)/\operatorname{Var}(T)$ reported in Table~\ref{tab:frontier} play an analogous role here: both are attempts to summarise, in a single number, how much of the panel's row count actually functions as identifying variation for $\theta$. Neither is derived as an effective-sample-size statistic in the formal sense of that literature, and \S\ref{sec:frontier} is explicit that $V$ is a heuristic index rather than a proven information measure; the connection is offered to situate the exercise, not to import its asymptotic guarantees.

\paragraph{Variance components across units.} To use multiple realised price histories, we adapt a standard random-effects variance-component estimator from meta-analysis \citep{paule1982consensus, dersimonian1986meta, hartung2001refined, borenstein2009introduction}. The Paule--Mandel estimator is not a mathematical contribution of the paper, and the general idea of pooling several independently randomised replications with a random-effects model is not new either: it is the logic behind treating a set of parallel randomized experiments as draws from a common process with unit-specific displacement, as in \citet{rubin1981estimation}. What we add is narrower and specific to the present setting: a characterisation of \emph{when} that pooling identifies design uncertainty and \emph{when} it does not. Its role is to estimate excess between-unit dispersion under a working model in which units provide approximately independent design realisations, within-unit variances are adequately measured, and the design-specific displacement is exchangeable with mean zero. When the true unit-level elasticities are common, this excess dispersion can be interpreted as design uncertainty under those assumptions, and \S\ref{sec:hierarchical}--\S\ref{sec:level} quantify the resulting coverage gain in the simulated grid. When true elasticities differ, the between-unit component generally combines genuine heterogeneity with design uncertainty, so it no longer identifies the latter without additional structure; \S\ref{sec:limitations} returns to this failure mode and to what would be needed---repeated designs per unit, or an external restriction on heterogeneity---to separate the two components in practice.

\paragraph{Relation to pricing and inference practice.} Limited and endogenous price variation is a long-standing problem in empirical demand estimation \citep{villas1999endogeneity, bijmolt2005new}, and recent pricing work emphasizes the value of controlled variation for learning demand and treatment heterogeneity \citep{ferreira2016analytics, dube2023personalized}. Our contribution is narrower: we use the simulation environment to distinguish additional rows observed under essentially the same price path from additional, weakly dependent price trajectories, and to quantify how that distinction affects repeated-design error. The centring issue is also related to bias-aware inference \citep{armstrong2018simple}, although the displacement studied here is generated by the realised pricing design rather than by a generic smoothing-bias bound. Multiway clustering and related covariance corrections remain relevant for dependence within the observed panel \citep{cameron2011robust, chiang2022multiway}; they answer a different question from learning the distribution of errors across unrealised pricing designs.

\paragraph{Scope of the evidence.} Proposition~\ref{prop:agg} below is an algebraic variance identity. All quantitative magnitudes---including the $97.6\%$ decomposition, the coverage results, the estimated exponent $-0.271$, and the variance-component performance---are simulation results from the stated DGP and grid, with one explicit exception: the real-data application of Section~\ref{sec:realdata}, labelled as such wherever it appears. We do not interpret the exponent as a general rate, the baseline variance share as a population constant, or the failure of the eight evaluated intervals as an impossibility theorem. The simulations are intended to isolate a mechanism and to show conditions under which data design, rather than row count alone, materially changes inference.

\section{Setup}\label{sec:setup}

\subsection{The estimand}

Let $i$ index product units, $r$ regions and $t$ weeks. The outcome is
$Y_{irt} = \log Q_{irt}$, log volume; the treatment is $T_{irt} = \log
P^{\text{si}}_{irt}$, the log list price the manufacturer sets. We work with the
partially linear model \citep{robinson1988root}
\begin{equation}
Y = \theta\, T + g(W) + \varepsilon, \qquad
T = m(W) + \eta, \qquad
\mathbb{E}[\varepsilon \mid W,\eta]=0,\ \ \mathbb{E}[\eta\mid W]=0,
\label{eq:plr}
\end{equation}
with $W$ a vector of observed confounders (promotional flag and intensity, log
competitor price, weather, holidays, working days, week of year, log inflation).
The parameter $\theta$ is estimated with a cross-fitted residual-on-residual procedure motivated by Double Machine Learning \citep{chernozhukov2018double}. Nuisance functions $\hat\ell = \hat{\mathbb{E}}[Y\mid W]$ and $\hat m = \hat{\mathbb{E}}[T \mid W]$ are fit on $K=5$ contiguous temporal blocks, with all regional observations of a week assigned to the same block. The implementation aggregates fold-specific coefficients by their median, as in \eqref{eq:dml}; because this is not the canonical pooled DML2 formula, all claims below should be read as applying to the estimator actually implemented here rather than to DML estimators as a class:
\begin{equation}
\hat\theta \;=\; \operatorname*{median}_{k}
\frac{\sum_{i \in I_k}\tilde T_i \tilde Y_i}{\sum_{i \in I_k}\tilde T_i^2},
\qquad
\tilde Y_i = Y_i - \hat\ell^{(-k)}(W_i),\quad
\tilde T_i = T_i - \hat m^{(-k)}(W_i).
\label{eq:dml}
\end{equation}

\begin{remark}[the estimand under split pricing]\label{rem:split}
In the channel that motivates this work the manufacturer sets the list price
while a shopkeeper sets the shelf price, so volume responds to a price the
manufacturer does not control. Writing $\rho = \partial \log P^{\text{so}} /
\partial \log P^{\text{si}}$ for the pass-through and $\varepsilon_c$ for the
consumer elasticity against the shelf price, the quantity identified by
\eqref{eq:dml} is the convolution $\theta = \varepsilon_c\,\rho$ and not
$\varepsilon_c$. Everything below concerns $\theta$; the separation of the two
is a distinct problem and is not our subject. We note the point only so that the
ground truth against which we measure is unambiguous: it is $\varepsilon_c \rho$
as realised in each generated panel, read from the generator rather than from its
configuration.
\end{remark}

\subsection{The generator}\label{sec:generator}

Panels are drawn from a data-generating process described in
Appendix~\ref{app:dgp}. Volume is generated jointly with prices, costs,
promotions, calendar effects, weather and an inflation index, so that
confounding is present by construction. The baseline configuration is $W=120$
weeks, six regions, list prices moving ten times per series by $3.0$--$7.5\%$,
pass-through $\rho = 0.85$, a cross-price elasticity of $+0.30$ on the log
competitor price, and a demand shock of standard deviation $0.05$ in logs. 

To ensure that this data-generating process reflects realistic market dynamics rather than a pathological edge case, these parameters are calibrated against the empirical moments in \citet{nakamura2008five}. In US microdata, regular price changes exhibit a median monthly frequency of $9$--$12\%$---corresponding to $8$--$11$ months of median duration, or roughly $2.5$--$3.3$ price movements over a $120$-week horizon. The median absolute size of regular price adjustments is $8.5\%$, whereas temporary sales are more than three times larger at a median of $29.5\%$. The baseline simulation regime directly mirrors these benchmark frequencies and magnitudes.

As a robustness check reported alongside Table~\ref{tab:decomp} and Equation~\eqref{eq:scaling}, we also re-parameterise the move-generating step directly from the Nakamura--Steinsson monthly frequency (about $12\%$) rather than from a fixed count of ten moves, and re-run the corresponding experiments with an independently coded implementation of the generator and estimator. This alternative calibration changes the realised twelve designs and, with them, the specific decomposition and exponent obtained; it does not change the qualitative pattern, and we report both where relevant so that the headline figures are not mistaken for fixed constants of the DGP.

A \emph{design} is a realisation of everything except the demand shock; a
\emph{replication} of a design is a fresh draw of that shock alone. This
separation is what makes Section~\ref{sec:decomposition} possible and is
implemented by giving the shock its own random stream.

Two learners recur throughout. \textsf{gbr} is a gradient-boosted regressor
(depth $3$, $200$ trees, learning rate $0.05$), the specification most common in
applied DML. \textsf{sieve} is a ridge on an explicit basis: the confounders, a
quadratic time trend and three annual harmonic pairs. They differ sharply in
bias and variance, and the contrast turns out to matter for the aggregation
result of Section~\ref{sec:aggregation}.

\section{Within- and between-design uncertainty}\label{sec:decomposition}

\begin{definition}[within- and between-design error dispersion]
Let $D$ denote a design, $u$ a shock realisation, and $\theta(D)$ the design-specific ground truth used by the generator. Define the estimation error
\begin{equation}
e(D,u)=\hat\theta(D,u)-\theta(D),
\end{equation}
and its design-specific conditional mean $b(D)=\mathbb{E}_u[e(D,u)\mid D]$. The \emph{within-design} variance is
\begin{equation}
\sigma_w^2=\mathbb{E}_D\!\left[\operatorname{Var}_u(e(D,u)\mid D)\right],
\end{equation}
and the \emph{between-design centring variance} is
\begin{equation}
\sigma_b^2=\operatorname{Var}_D\!\left(b(D)\right).
\end{equation}
By the law of total variance,
\begin{equation}
\operatorname{Var}_{D,u}(e(D,u))=\sigma_w^2+\sigma_b^2.
\end{equation}
If $\bar b=\mathbb{E}_D[b(D)]$ is nonzero, the corresponding mean-squared error is $\sigma_w^2+\sigma_b^2+\bar b^2$.
\end{definition}

\begin{definition}[conditional and repeated-design coverage]
Let $CI_{1-\alpha}(D,u)$ denote an interval computed from one realised panel. Its \emph{conditional coverage} given $D$ is
\begin{equation}
\Pr_u\!\left(\theta(D)\in CI_{1-\alpha}(D,u)\mid D\right).
\end{equation}
Its \emph{repeated-design coverage} under the simulation DGP is
\begin{equation}
\Pr_{D,u}\!\left(\theta(D)\in CI_{1-\alpha}(D,u)\right)
=\mathbb{E}_D\!\left[\Pr_u\!\left(\theta(D)\in CI_{1-\alpha}(D,u)\mid D\right)\right].
\end{equation}
The equality is a probability identity; the substantive distinction is which source of randomness is treated as relevant to the inferential question.
\end{definition}

A within-panel procedure is constructed from one realised design. Depending on the resampling scheme and dependence structure, it may approximate aspects of the conditional distribution more or less well, but it does not by itself provide nonparametric information about the distribution of $b(D)$ over price trajectories that were not realised. We therefore do not equate the bootstrap standard error mechanically with $\sigma_w$. Instead, we measure $\sigma_w$ and $\sigma_b$ directly from the generator. We draw $12$ designs and $40$ shock replications within each, estimate $\theta$ on all $480$ panels, and compute the error decomposition relative to each design's ground truth.

\begin{table}[t]
\centering
\caption{Decomposition of the dispersion of the estimation error. Twelve designs, forty
shock replications each ($n=480$). $\sigma_w$ is measured within design,
$\sigma_b$ is the standard deviation of the design-specific conditional mean error $b(D)$; $\widehat{\text{se}}$ is the weekly moving-block bootstrap standard error from the baseline specification. Coverage is averaged over the design-specific conditional coverage probabilities and is measured against each design's own ground truth, at a nominal $0.95$.}
\label{tab:decomp}
\begin{tabular}{lrrrrrr}
\toprule
Learner & Bias & $\sigma_w$ & $\sigma_b$ & $\widehat{\text{se}}$
        & $\sigma_b^2/(\sigma_w^2{+}\sigma_b^2)$ & Coverage \\
\midrule
\textsf{gbr}   & $0.150$ & $0.0756$ & $0.4815$ & $0.1595$ & $97.6\%$ & $0.565$ \\
\textsf{sieve} & $0.016$ & $0.0589$ & $0.7187$ & $0.1553$ & $99.3\%$ & $0.242$ \\
\bottomrule
\end{tabular}
\end{table}

Table~\ref{tab:decomp} reports the result. Three features deserve comment.

\paragraph{Between-design centring dispersion is large in the baseline DGP.} It accounts for $97.6\%$ of the variance of the estimation error for \textsf{gbr} and $99.3\%$ for \textsf{sieve}. Conditional on a design, the estimator is relatively concentrated---$\sigma_w=0.0756$ for \textsf{gbr}---but its conditional mean error varies substantially across designs. These percentages characterize the simulated generator and should not be read as universal variance shares for short pricing panels.

\paragraph{The exact share is itself imprecisely estimated.} $\sigma_b$ is computed from only twelve designs, so the reported shares carry non-trivial sampling uncertainty of their own. This is not a rhetorical caveat: using the raw per-design, per-shock error cache from the recalibrated run described next, we resample the twelve designs with replacement (a cluster bootstrap over designs, $5{,}000$ replications, recomputing $\sigma_w$, $\sigma_b$, and the share on each resample) and obtain a $95\%$ interval of $[64.3\%,96.2\%]$ for \textsf{gbr} and $[84.3\%,95.5\%]$ for \textsf{sieve}; leaving out one of the twelve designs at a time (a jackknife, which is less conservative because it never removes more than one design at once) narrows this to $[89.4\%,94.5\%]$ and $[89.5\%,93.6\%]$ respectively. An independent re-implementation of the generator and estimator, run at the same replication counts ($12\times40$) but with an unrelated random stream and the design-frequency parameterisation of \S\ref{sec:generator} recalibrated to the monthly price-change frequency reported by \citet{nakamura2008five} (about $12\%$) rather than a fixed count of ten moves, is the run that produced this cache; its point shares are $93.3\%$ for \textsf{gbr} and $92.7\%$ for \textsf{sieve}, with $\sigma_b=0.273$ and $0.217$. The original run reported in Table~\ref{tab:decomp} returns higher point shares, $97.6\%$ for \textsf{gbr} and $99.3\%$ for \textsf{sieve}: the first sits at the upper edge of the cluster-bootstrap interval above, and the second sits just outside it. The gap between the two runs is therefore not fully attributable to the sampling noise of drawing twelve designs; part of it reflects the change in move-frequency parameterisation itself, and the bootstrap lets us say so quantitatively rather than by assertion. What survives both the recalibration and the resampling is the qualitative claim the paper relies on: even the most pessimistic endpoint of the widest interval, $64\%$, still places most of the total error variance between designs rather than within them. The second or third significant digit of the percentage is not something twelve designs can pin down, and should not be read as a fixed constant of the DGP.

\paragraph{The bootstrap standard error is not an estimate of the full decomposition.} For \textsf{gbr}, the mean bootstrap standard error, $0.1595$, is about $2.1$ times the measured within-design standard deviation and about one third of $\sigma_b$. Thus it is not accurate to describe this bootstrap as simply estimating $\sigma_w$ in the finite sample considered here. The more limited conclusion is that a procedure built from one realised panel does not reproduce the across-design distribution measured by repeated generation of price trajectories.

\paragraph{The simulations point to a centring component.} The observed coverage shortfall is difficult to reconcile with interval width alone. For \textsf{gbr}, a typical half-width of about $0.305$ is large relative to $\sigma_w=0.0756$, yet mean conditional coverage across designs is $0.565$. In the generator, the design-specific conditional mean error $b(D)$ varies substantially across designs. A rough calculation based on the dispersion of $b(D)$ yields coverage of the same order as the simulation. We interpret this as evidence that design-dependent centring is an important mechanism in this DGP, not as a proof that every resampling method fails whenever treatment variation is sparse.

\begin{remark}[which target is the right one]
For a firm interpreting one realised historical trajectory, conditional coverage is a natural target. For evaluating a procedure that will be reused across future pricing histories, repeated-design coverage is also informative. Neither target dominates by definition; the relevant one depends on the decision problem. The simulations report both so that a conditional inferential question is not inadvertently replaced by an unconditional one.
\end{remark}

\section{No evaluated within-panel construction achieves nominal coverage}\label{sec:eight}

To assess whether familiar within-panel procedures are sufficient in this DGP, we compare eight constructions on identical fits over $200$ independently drawn panels: the baseline moving-block bootstrap over weeks; the same at $600$ draws; a hierarchical block bootstrap that also
resamples cells within replicated weeks; the analytic i.i.d.\ standard error;
cluster-robust by week; multiway cluster-robust by week and region
\citep{cameron2011robust, chiang2022multiway}; a Newey--West long-run variance
on the weekly score \citep{newey1987simple}; and a wild cluster bootstrap-$t$
with Rademacher weights \citep{cameron2008bootstrap}.

\begin{table}[t]
\centering
\caption{Eight interval constructions, identical point estimator and identical
fits, over $200$ panels. Nominal level $0.95$. \emph{Reference repeated-design width} is $3.92\,\hat\sigma$ with $\hat\sigma$ the realised across-panel standard deviation of the estimation error. This is a descriptive benchmark, not a confidence interval with a formal finite-sample coverage guarantee.}
\label{tab:eight}
\begin{tabular}{lrrrr}
\toprule
& \multicolumn{2}{c}{\textsf{gbr}} & \multicolumn{2}{c}{\textsf{sieve}} \\
\cmidrule(lr){2-3}\cmidrule(lr){4-5}
Construction & Coverage & Width & Coverage & Width \\
\midrule
Multiway cluster (week $\times$ region) & $0.730$ & $1.062$ & $0.660$ & $1.167$ \\
Block bootstrap, cells resampled       & $0.595$ & $0.712$ & $0.435$ & $0.723$ \\
Newey--West on the weekly score        & $0.580$ & $0.798$ & $0.230$ & $0.451$ \\
Block bootstrap, $B=600$               & $0.545$ & $0.591$ & $0.300$ & $0.581$ \\
Block bootstrap, $B=250$ (baseline)  & $0.525$ & $0.597$ & $0.310$ & $0.566$ \\
Wild cluster bootstrap-$t$             & $0.350$ & $0.451$ & $0.105$ & $0.258$ \\
Cluster-robust by week                 & $0.345$ & $0.445$ & $0.105$ & $0.254$ \\
Analytic i.i.d.                        & $0.285$ & $0.338$ & $0.155$ & $0.308$ \\
\midrule
\emph{Reference repeated-design width} $3.92\,\hat\sigma$ & \multicolumn{2}{c}{$1.680$}
                                       & \multicolumn{2}{c}{$2.212$} \\
\bottomrule
\end{tabular}
\end{table}

None of the eight evaluated constructions reaches $0.95$ coverage in this experiment (Table~\ref{tab:eight}). Coverage is generally higher for wider intervals, and multiway clustering has the highest measured coverage for both learners among the procedures considered. This ranking should not be read as an impossibility result for all within-panel methods; it is a finite simulation comparison of these eight constructions. The table also emphasizes that the empirical shortfall is not resolved simply by choosing the widest evaluated interval.

For the decision exercise used in this paper, we pre-specify an application-specific width threshold of $W\le 0.6$. This threshold is not a general standard for pricing studies. Under that criterion, the higher-coverage intervals in Table~\ref{tab:eight} remain too wide to satisfy the paper's operational precision requirement. The resulting coverage--width trade-off motivates the across-unit variance-component exercise in Section~\ref{sec:hierarchical}.

\section{An empirical dispersion pattern over the simulation grid}\label{sec:frontier}

If design variance is what binds, the natural question is how much identifying
variation would be needed to make it small. We sweep the number of price moves
$n \in \{2,4,6,8,12,16\}$, their magnitude (four ranges from $0.4$--$0.9\%$ to
$3.0$--$7.5\%$), the degree of confounding with promotion
($\{0, 0.5, 0.9\}$) and the pass-through ($\{0, 0.30, 0.60, 0.85\}$): $288$ grid
points, each estimated with both learners and replicated eight times, for $576$
configurations and $4{,}608$ fits in total. At each point we measure the realised
dispersion of $\hat\theta$ across replications rather than its reported standard
error.

\begin{table}[t]
\centering
\caption{A selected column of the simulation grid: no confounding with
promotion, $\rho = 0.85$, moves averaging $5.25\%$, learner \textsf{gbr}. True
$\theta = -1.114$; eight replications per grid point, so $\sigma$ is itself
estimated with visible noise --- which is why the fit \eqref{eq:scaling} is taken
over the whole column rather than read off adjacent rows. $\operatorname{Var}
(\tilde T)/\operatorname{Var}(T)$ is the share of the log list price surviving
residualisation, i.e.\ the identifying variation the estimator actually uses. \textit{Filtered ratio} indicates the proportion of simulated configurations satisfying the pre-estimation identification criterion (a treatment coefficient of variation $CV > 0.03$ and at least four discrete price movements).}
\label{tab:frontier}
\begin{tabular}{rrrrr}
\toprule
Moves & Bias & $\sigma$ (design)
      & $\operatorname{Var}(\tilde T)/\operatorname{Var}(T)$ & Filtered ratio \\
\midrule
$2$  & $0.254$ & $0.681$ & $0.352$ & $0.000$ \\
$4$  & $0.243$ & $0.869$ & $0.407$ & $0.375$ \\
$6$  & $0.471$ & $0.429$ & $0.530$ & $0.688$ \\
$8$  & $0.290$ & $0.444$ & $0.491$ & $0.729$ \\
$12$ & $0.290$ & $0.394$ & $0.569$ & $0.896$ \\
$16$ & $0.145$ & $0.382$ & $0.547$ & $0.958$ \\
\bottomrule
\end{tabular}
\end{table}

Let $V=n\cdot\text{magnitude}^2$ summarize the amount and size of list-price movement. The index is a deliberately simple design summary; it is not asserted to be Fisher information or a sufficient statistic for identification. Over the $24$ points with no promotion confounding and full pass-through---six move counts $\times$ four magnitude ranges for learner \textsf{gbr}---a log--log regression gives
\begin{equation}
\hat\sigma_b \;=\; 0.182\;V^{-0.271},\qquad
\text{s.e.}(-0.271)=0.023,\qquad R^2=0.86.
\label{eq:scaling}
\end{equation}

Equation~\eqref{eq:scaling} is an empirical regularity of this simulation column. The exponent $-0.271$ should not be interpreted as a theoretical convergence rate for DML, for short panels, or for observational pricing designs in general. A $-1/2$ slope can serve as a heuristic reference in stylized settings where information grows proportionally with a sum of squared treatment movements, but $V$ is not shown here to satisfy that information equivalence. The comparison is therefore descriptive rather than an asymptotic test.

\begin{figure}[t]
\centering
\includegraphics[width=0.85\textwidth]{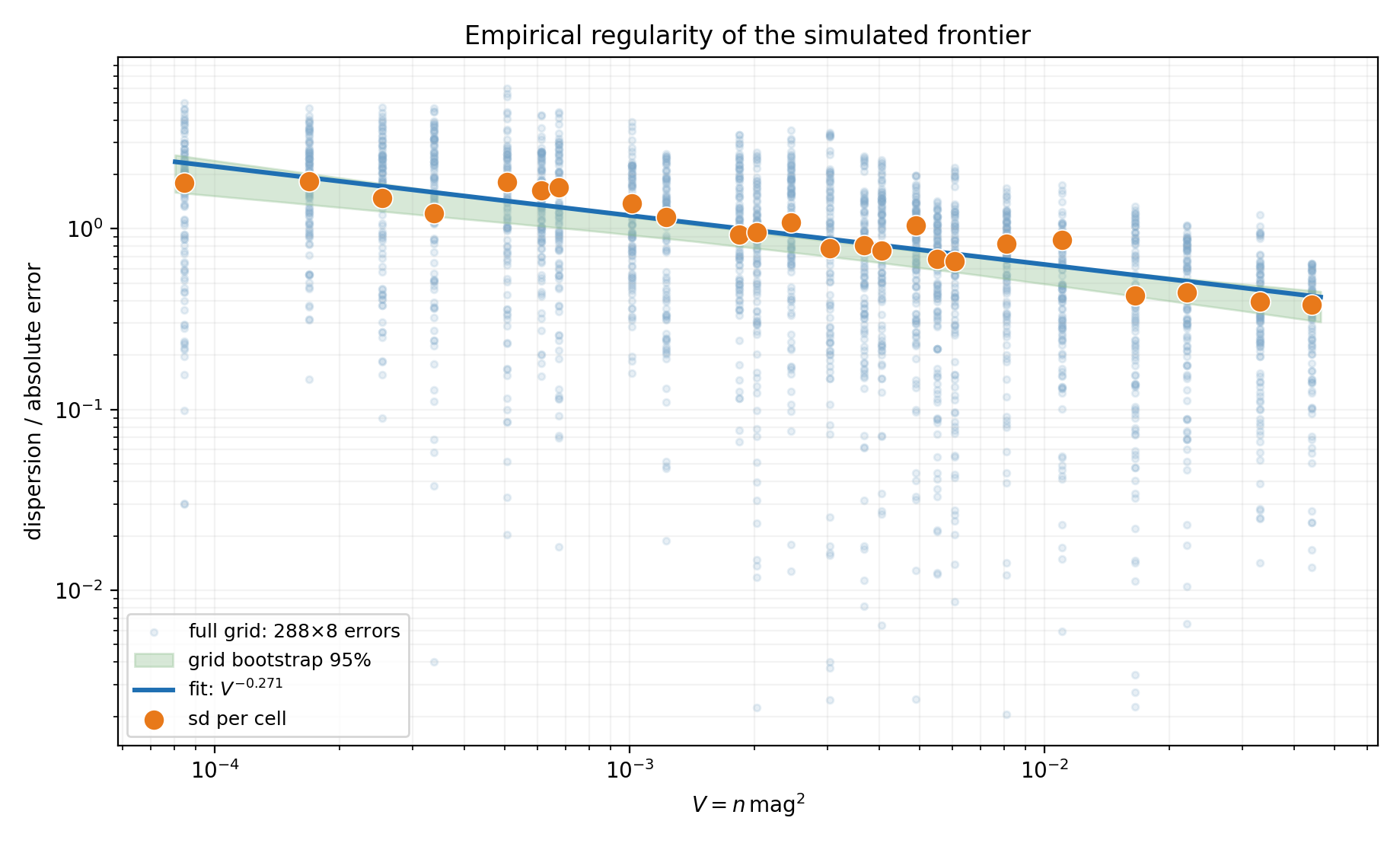}
\caption{The fitted relation \eqref{eq:scaling} against the full simulation grid underlying it ($288$ configurations $\times\,8$ replications $=2{,}304$ points, learner \textsf{gbr}). Light points are individual replication errors, plotted at the $V$ of their own configuration regardless of promotion confounding or pass-through; solid orange markers are the per-cell standard deviation over the $24$ cells with no promotion confounding and full pass-through, the same $24$ points used to fit Equation~\eqref{eq:scaling}; the shaded band is the $95\%$ grid bootstrap region for the fit described above. The fitted line reproduces $\hat\sigma_b=0.182\,V^{-0.271}$ exactly.}
\label{fig:frontier}
\end{figure}

\paragraph{Sampling uncertainty of the exponent.} The reported standard error already signals that $-0.271$ is estimated from a small column of the grid---$24$ points at eight replications each. A percentile bootstrap on the raw per-replication errors underlying this column (resampling the eight replications within each of the $24$ cells with replacement, $5{,}000$ refits of the log--log regression) gives a $95\%$ interval of $[-0.33,-0.21]$ for the exponent and $[0.11,0.24]$ for the coefficient $\alpha$. An independent re-implementation of the generator and estimator, run on the same grid and replication counts but with an unrelated random stream and the Nakamura-calibrated move-frequency parameterisation described above, returned $\hat\sigma_b=0.141\,V^{-0.192}$ (s.e.\ on the exponent $0.032$, $R^2=0.63$): a point estimate that sits just outside this interval rather than inside it. As with the between-design share in \S\ref{sec:decomposition}, the gap is not fully attributable to the sampling noise of an eight-replication-per-cell bootstrap; part of it reflects the change in move-frequency parameterisation itself. What the bootstrap does establish is that the exponent is not stable enough across reseeding, let alone across DGP re-parameterisation, to be quoted to three decimal places, even though the qualitative pattern---dispersion falling slowly in $V$---replicates in both runs. Figure~\ref{fig:frontier} shows why: plotted against the full grid of $288\times8$ individual errors underlying Equation~\eqref{eq:scaling} itself, rather than against the $24$ per-cell summary points alone, the fitted line and its bootstrap band sit inside a cloud spanning several orders of magnitude at every value of $V$. The per-cell standard deviations used to fit \eqref{eq:scaling} are a thin summary of that cloud, and the figure is meant to keep that fact visible rather than let the single fitted exponent stand in for it.

Within the simulated grid, larger $V$ is associated with lower across-design dispersion, but the decline is slow enough that the explored configurations retain substantial error dispersion. Of the $576$ learner--configuration combinations, none simultaneously attains absolute bias below $0.15$ and design dispersion below $0.20$. This is a statement about the enumerated grid and its finite replication count, not an impossibility result outside the grid. The best cells should likewise be interpreted cautiously because each grid point has only eight replications.

\begin{remark}[on extrapolating \eqref{eq:scaling}]\label{rem:extrap}
If one mechanically extrapolates the fitted relation, halving $\sigma_b$ corresponds to multiplying $V$ by approximately $2^{1/0.271}\approx13$. Such calculations extend beyond the simulated support and are included only to convey the curvature of the fitted relationship. They are not forecasts of how a real category would respond to thirteen times more price variation.
\end{remark}

\paragraph{Implication for precision calculations.} On this grid, precision measures based on the reported within-panel standard errors can be much smaller than the realised repeated-design error dispersion. The median ratio of repeated-design dispersion to the reported standard-error scale is $2.26$ for \textsf{gbr} and $8.03$ for \textsf{sieve}. These ratios are simulation diagnostics, not generic correction factors for minimum detectable effects.

\section{Common versus independent price paths under aggregation}\label{sec:aggregation}

The key distinction is not panel width by itself but the covariance structure of the design-specific estimation errors contributed by the added units.

\begin{proposition}[variance of averaged design-specific centring errors]
\label{prop:agg}
Let $b_j$ be the design-specific conditional mean error for unit $j$, and let $\bar b_k=k^{-1}\sum_{j=1}^k b_j$. Then
\begin{equation}
\operatorname{Var}(\bar b_k)
=\frac{1}{k^2}\left(\sum_{j=1}^k\operatorname{Var}(b_j)
+2\sum_{j<\ell}\operatorname{Cov}(b_j,b_\ell)\right).
\label{eq:aggvar}
\end{equation}
If $\operatorname{Var}(b_j)=\sigma_b^2$ and all pairwise correlations equal $\rho$, then
\begin{equation}
\operatorname{Var}(\bar b_k)=\sigma_b^2\frac{1+(k-1)\rho}{k}.
\label{eq:equicorr}
\end{equation}
Hence $\rho=0$ gives a standard deviation of $\sigma_b/\sqrt{k}$, while $\rho=1$ gives no reduction. Aggregation does not remove a nonzero common mean bias $\mathbb{E}[b_j]$.
\end{proposition}

The proposition is an algebraic identity, not a claim that shared prices automatically imply $\rho=1$ or that distinct price paths automatically imply $\rho=0$. In the simulation, regions exposed to the same national list-price trajectory inherit a large common design component, whereas product units are generated with distinct price trajectories and approximately weak cross-unit dependence in that component. The empirical exercise below asks whether the observed covariance structure is close enough to the independent benchmark to produce the $\sqrt{k}$ pattern.

\begin{table}[t]
\centering
\caption{Estimating the same object at three levels of aggregation, over $200$
panels of eight product units in two brands of one category. Residualisation is
always per unit. The Frisch--Waugh--Lovell logic \citep{lovell1963seasonal} relates the pooled residual-on-residual coefficient to a specification with unit effects; the exercise is designed so that aggregation combines information from multiple unit-specific price histories. \emph{Units} is the number of units pooled per estimate.
$\sigma$ (design) is the dispersion across replications, and \emph{Width} is the mean width of the baseline block-bootstrap interval at that level.}
\label{tab:levels}
\begin{tabular}{llrrrrrrr}
\toprule
Learner & Level & Units & Bias & $\sigma$ (design) & RMSE & Coverage & Width
        & $\sigma$ ratio \\
\midrule
\textsf{gbr}   & unit     & $1$ & $0.315$ & $0.4603$ & $0.559$ & $0.474$ & $0.693$ & --- \\
\textsf{gbr}   & brand    & $4$ & $0.293$ & $0.2219$ & $0.368$ & $0.370$ & $0.414$ & $2.07$ \\
\textsf{gbr}   & category & $8$ & $0.291$ & $0.1538$ & $0.327$ & $0.280$ & $0.364$ & $2.99$ \\
\addlinespace
\textsf{sieve} & unit     & $1$ & $0.021$ & $0.4737$ & $0.475$ & $0.282$ & $0.422$ & --- \\
\textsf{sieve} & brand    & $4$ & $0.021$ & $0.2267$ & $0.227$ & $0.282$ & $0.191$ & $2.09$ \\
\textsf{sieve} & category & $8$ & $0.022$ & $0.1585$ & $0.157$ & $0.240$ & $0.134$ & $2.99$ \\
\midrule
\multicolumn{6}{l}{\emph{Predicted by Proposition~\ref{prop:agg}}}
& \multicolumn{3}{r}{$\sqrt{4}=2.00$, \ $\sqrt{8}=2.83$} \\
\bottomrule
\end{tabular}
\end{table}

Table~\ref{tab:levels} is close to the independent-design benchmark in this simulation: the observed reduction factors are $2.07$ and $2.09$ against $\sqrt{4}=2.00$, and $2.99$ against $\sqrt{8}=2.83$, for both learners. This is evidence that the simulated cross-unit design-specific errors have low enough covariance for the $\sqrt{k}$ approximation to be useful here; it is not evidence that product-level price histories are independent in empirical applications.

Two further readings of the table matter. First, aggregation improves RMSE substantially in the reported simulations, but the baseline interval does not gain coverage because its width contracts along with the empirical error dispersion. Second, the two learners respond differently. The \textsf{gbr} mean bias remains near $0.30$ across aggregation levels, which is consistent with a persistent common centring component; its coverage falls as the interval tightens. The \textsf{sieve} mean bias remains near $0.02$, so the reduction in weakly correlated dispersion translates more directly into lower RMSE. These are empirical patterns of the simulated portfolio, not generic properties of boosted trees or sieve learners.

\begin{table}[t]
\centering
\caption{Six nuisance learners at unit level, $120$ panels. \emph{Low-frequency
share} is the fraction of the residual's spectral mass below one twentieth of the
series length. No learner minimises bias, maximises coverage and minimises RMSE
simultaneously.}
\label{tab:learners}
\begin{tabular}{lrrrr}
\toprule
Learner & Bias & RMSE & Coverage & Low-frequency share \\
\midrule
\textsf{lasso}  & $-0.007$ & $0.652$ & $0.258$ & $0.508$ \\
\textsf{ridge}  & $-0.008$ & $0.651$ & $0.250$ & $0.510$ \\
\textsf{sieve}  & $-0.009$ & $0.652$ & $0.258$ & $0.505$ \\
\textsf{rbf}    & $ 0.055$ & $0.534$ & $0.533$ & $0.277$ \\
\textsf{histgb} & $ 0.153$ & $0.517$ & $0.383$ & $0.128$ \\
\textsf{gbr}    & $ 0.176$ & $0.501$ & $0.383$ & $0.144$ \\
\bottomrule
\end{tabular}
\end{table}

Table~\ref{tab:learners} places this in context. At unit level the choice of
nuisance learner is a genuine trade-off: the regularised learners are nearly
unbiased and have the worst RMSE and coverage, the boosted learners are the
reverse, and no learner wins on all three. The ranking therefore changes with the aggregation level in these simulations. This illustrates, rather than proves generally, that nuisance-learner comparisons can depend on the level at which the estimand is reported, because aggregation attenuates weakly correlated dispersion but not a persistent common mean bias.

\section{A variance-component interval estimated across realised designs}\label{sec:hierarchical}

A single realised price trajectory does not nonparametrically identify the distribution of $b(D)$ across counterfactual trajectories. Multiple units with independently or weakly dependently realised pricing designs can supply information about that distribution, but only under additional structure. We use the working model
\begin{equation}
\hat\theta_j=\theta_j+B_j+\varepsilon_j,\qquad
\mathbb{E}[B_j]=0,\quad \operatorname{Var}(B_j)=\tau_B^2,\quad
\mathbb{E}[\varepsilon_j]\approx0,\quad \operatorname{Var}(\varepsilon_j)\approx s_j^2,
\label{eq:random_bias}
\end{equation}
with $B_j$ exchangeable across units and approximately independent of the within-unit error $\varepsilon_j$. The interpretation of $\tau_B^2$ as design variance additionally requires that the units' pricing trajectories act as independent or sufficiently weakly dependent design draws and that $s_j^2$ adequately represents the within-unit component.

Let $\hat\theta_j$ and $s_j^2$ be the estimate and bootstrap variance of unit $j$, $j=1,\dots,J$. We estimate an excess between-unit variance with the Paule--Mandel moment equation \citep{paule1982consensus}
\begin{equation}
\sum_j\frac{(\hat\theta_j-\hat\mu)^2}{s_j^2+\tau^2}=J-1,
\qquad
\hat\mu=\frac{\sum_j(s_j^2+\tau^2)^{-1}\hat\theta_j}{\sum_j(s_j^2+\tau^2)^{-1}}.
\label{eq:pm}
\end{equation}
Paule--Mandel is a standard between-study variance estimator; the contribution here is its use as a working estimator of excess dispersion across independently realised pricing designs, not a new variance-component method. We compare the bootstrap percentile interval; a conjugate normal--normal posterior interval based on a leave-one-out branch prior; a variance-augmented interval $\hat\theta_j\pm q\sqrt{s_j^2+\hat\tau^2}$; and its posterior-centred analogue $\hat\theta_j^{\mathrm{post}}\pm q\sqrt{v_j^{\mathrm{post}}+\hat\tau^2}$, where $q=z_{1-\alpha/2}$.

The homogeneous and heterogeneous scenarios have different interpretations. In the homogeneous scenario all eight units share the same true $\theta$ and have independently generated price paths. Under model \eqref{eq:random_bias}, excess between-unit dispersion can then be attributed to the design component up to estimation error in $s_j^2$ and $\hat\tau^2$. In the heterogeneous scenario the true $\theta_j$ differ, so the same between-unit component generally combines true effect heterogeneity and design-specific displacement. It should therefore be viewed as a conservative mixture rather than as a clean estimate of design variance.

\begin{table}[t]
\centering
\caption{Four intervals on identical fits, $200$ panels of eight units per
scenario ($n=1{,}600$ unit-panels). Nominal level $0.95$. In the homogeneous
scenario the eight units share the same true $\theta$, so $\hat\tau$ is interpretable as excess design-related dispersion under the working-model assumptions described in the text.}
\label{tab:hier}
\begin{tabular}{llrr}
\toprule
Scenario & Interval & Coverage & Width \\
\midrule
Homogeneous & Posterior-centred variance-augmented $\sqrt{v^{\text{post}}+\hat\tau^2}$ & $0.931$ & $1.480$ \\
Homogeneous & Variance-augmented $\sqrt{s^2+\hat\tau^2}$                      & $0.906$ & $1.516$ \\
Homogeneous & Bootstrap percentile                                     & $0.469$ & $0.549$ \\
Homogeneous & Posterior $\sqrt{v^{\text{post}}}$                       & $0.426$ & $0.501$ \\
\addlinespace
Heterogeneous & Posterior-centred variance-augmented                    & $0.943$ & $1.623$ \\
Heterogeneous & Variance-augmented                                     & $0.931$ & $1.650$ \\
Heterogeneous & Bootstrap percentile                                   & $0.475$ & $0.539$ \\
Heterogeneous & Posterior                                              & $0.426$ & $0.502$ \\
\bottomrule
\end{tabular}
\end{table}

Table~\ref{tab:hier} reports the outcome. In the homogeneous scenario, adding the estimated between-unit component raises empirical coverage from $0.469$ for the bootstrap percentile interval to $0.931$ for the posterior-centred variance-augmented interval. This is a substantial movement toward the nominal $0.95$ level, but it is not exact nominal coverage. The estimated $\hat\tau=0.372$ is much larger than the mean bootstrap standard error of $0.144$, which is consistent with substantial excess dispersion across independently generated unit-level designs in this simulation.

\paragraph{Shrinkage alone does not address the simulated centring dispersion.} The posterior interval has coverage $0.426$, below the bootstrap percentile interval. In this DGP, shrinkage reduces the conditional variance while leaving design-specific displacement largely unmodelled. The result is consistent with the centring mechanism documented in Section~\ref{sec:decomposition}; it should not be generalized to hierarchical Bayes procedures that explicitly model the design process or bias component.

\paragraph{The coverage improvement is purchased with width.} The two variance-augmented constructions are $2.70$ and $2.76$ times as wide as the bootstrap in the homogeneous simulation, at widths of $1.480$ and $1.516$. Under the paper's pre-specified operational threshold, these intervals are too wide to support the intended unit-level pricing decision. The result illustrates the trade-off in this DGP: representing excess across-design dispersion improves coverage but can materially reduce decision precision.

\section{The variance-component interval at higher aggregation levels}\label{sec:level}

Sections~\ref{sec:aggregation} and \ref{sec:hierarchical} suggest a natural combined exercise. In the simulated portfolio, aggregation is close to the independent-design $\sqrt{k}$ benchmark, while the variance-component construction produces a wider interval at unit level. Combining them is
therefore the natural question, and it is the one on which the practical value
of the whole argument turns: does the variance-component interval, evaluated at a higher
level of aggregation, become narrow enough to support a decision?

Arithmetic from Tables~\ref{tab:levels} and \ref{tab:hier} suggests it should.
At category level the design dispersion is $0.1585$ and the bootstrap standard
error $0.034$, so the variance-augmented interval would have width $2 \times 1.96 \times \sqrt{0.034^2 + 0.1585^2} = 0.636$, against the $1.516$ that the same construction attains at unit level in Table~\ref{tab:hier} --- moving from uninformative toward the target precision bound. The arithmetic is not decisive,
however, because $\hat\tau$ is re-estimated among the units of whatever level is
being published, and its magnitude at that level is not the design dispersion of
Table~\ref{tab:levels}. We therefore run the construction directly, on a
portfolio of $24$ units --- six categories, two brands each, two units per brand
--- estimating $\theta$ at each of the three levels and solving \eqref{eq:pm}
between the units of that level: $24$ units at unit level, $12$ at brand, $6$ at
category, over $100$ replications per scenario.

\begin{table}[t]
\centering
\caption{The variance-augmented interval $\hat\theta \pm q\sqrt{s^2+\hat\tau^2}$ at three
levels of aggregation, $100$ replications per scenario on a portfolio of $24$
units ($6$ categories $\times$ $2$ brands $\times$ $2$ units). $\hat\tau$ is
estimated between the units of each level, so its $\sqrt{k}$ column is the
prediction of Proposition~\ref{prop:agg} applied to $k$ = units pooled per
estimate. \emph{Ratio to threshold} indicates interval width expressed as a multiple 
of the $0.6$ operational precision bound; no combination falls below $1.0$.}

\label{tab:nivel}
\begin{tabular}{llrrrrrr}
\toprule
Scenario & Learner & Level & $k$ & $\hat\tau$ & $\sqrt{k}$ pred.
          & Coverage & Rel. Width ($W/0.6$) \\
\midrule
Homogeneous & \textsf{gbr}   & unit     & $1$ & $0.4052$ & ---      & $0.899$ & $1.725$ \ ($2.88$) \\
Homogeneous & \textsf{gbr}   & brand    & $2$ & $0.2779$ & $0.2865$ & $0.830$ & $1.191$ \ ($1.99$) \\
Homogeneous & \textsf{gbr}   & category & $4$ & $0.1908$ & $0.2026$ & $0.665$ & $0.829$ \ ($1.38$) \\
\addlinespace
Homogeneous & \textsf{sieve} & unit     & $1$ & $0.4569$ & ---      & $0.953$ & $1.816$ \ ($3.03$) \\
Homogeneous & \textsf{sieve} & brand    & $2$ & $0.3104$ & $0.3231$ & $0.958$ & $1.221$ \ ($2.04$) \\
Homogeneous & \textsf{sieve} & category & $4$ & $0.2268$ & $0.2285$ & $0.962$ & $0.861$ \ ($1.43$) \\
\addlinespace
Heterogeneous & \textsf{gbr}   & unit     & $1$ & $0.4471$ & ---      & $0.938$ & $1.877$ \ ($3.13$) \\
Heterogeneous & \textsf{gbr}   & brand    & $2$ & $0.3449$ & $0.3161$ & $0.930$ & $1.432$ \ ($2.39$) \\
Heterogeneous & \textsf{gbr}   & category & $4$ & $0.2850$ & $0.2236$ & $0.893$ & $1.166$ \ ($1.94$) \\
\addlinespace
Heterogeneous & \textsf{sieve} & unit     & $1$ & $0.5039$ & ---      & $0.974$ & $2.003$ \ ($3.34$) \\
Heterogeneous & \textsf{sieve} & brand    & $2$ & $0.3832$ & $0.3563$ & $0.983$ & $1.506$ \ ($2.51$) \\
Heterogeneous & \textsf{sieve} & category & $4$ & $0.3249$ & $0.2520$ & $0.993$ & $1.250$ \ ($2.08$) \\
\bottomrule
\end{tabular}
\end{table}

For the evaluated portfolio and threshold, the result is negative (Table~\ref{tab:nivel}). Three readings are useful.

\paragraph{Coverage remains near or above nominal for one learner in this experiment.}
\textsf{sieve} covers $0.953$, $0.958$ and $0.962$ at the three levels of the
homogeneous scenario, and $0.974$ to $0.993$ in the heterogeneous one. Its
$\hat\tau$ declines close to the independent-design benchmark ($0.4569\to0.3104\to0.2268$ versus $0.3231$ and $0.2285$), providing a second simulation check of the covariance pattern in Section~\ref{sec:aggregation}. For \textsf{gbr}, the mean bias remains persistent as aggregation increases and coverage declines from $0.899$ to $0.665$ in the homogeneous scenario. In the heterogeneous scenario the decline is milder, while $\hat\tau$ also includes genuine effect heterogeneity.

\paragraph{The width remains above the pre-specified threshold.} The narrowest interval that covers
at all is \textsf{sieve} at category level, $0.861$, still $1.43$ times the
operational precision threshold. Of the twelve combinations none covers $0.90$ within a width of $0.6$; the closest approach in coverage, \textsf{sieve} at category
level in the heterogeneous scenario, covers $0.993$ at a width of $1.250$. The
arithmetic prediction of $0.636$ was optimistic for the reason anticipated
above: $\hat\tau$ among the six category-level units of this portfolio is
$0.2268$, not the $0.1585$ design dispersion of Table~\ref{tab:levels}, whose
category aggregate pools eight units rather than four.

\paragraph{In the heterogeneous scenario the interval is wider.} With truths that differ across categories, $\hat\tau$ falls more slowly
than $\sqrt{k}$ ($0.5039 \to 0.3832 \to 0.3249$ against $0.3563$ and $0.2520$),
because part of the between-unit component is genuine heterogeneity rather than design error. The resulting interval has coverage $0.993$ in the reported cell. Under the working model this is consistent with a conservative mixture of effect heterogeneity and design dispersion; it should not be interpreted as calibrated estimation of either component separately.

\begin{remark}[what would fit]
Extrapolating the $\sqrt{k}$ rate from the homogeneous \textsf{sieve} row, and
holding $s \approx 0.05$, a width of $0.6$ requires $\sqrt{s^2+\hat\tau^2} \le
0.153$, hence $\hat\tau \le 0.145$ and $k \approx (0.4569/0.145)^2 \approx 10$
units with independently realised price paths per published estimate. This
portfolio offers four. Whether an aggregate spanning roughly ten independently priced units remains aligned with the operational decision unit is application-specific. The extrapolation carries the caveat of
Remark~\ref{rem:extrap} --- it extends an observed rate one doubling beyond the
range measured.
\end{remark}

For the particular portfolio, learners, scenarios and width threshold studied here, no evaluated aggregation level simultaneously achieves the paper's coverage and precision criteria. This motivates examining the assignment mechanism itself, while stopping short of an impossibility claim for other estimators, portfolios or data-generating processes.

\paragraph{The point estimate, unlike the honest interval, can reach a usable precision.} Table~\ref{tab:levels} already shows that \textsf{sieve}'s root-mean-squared error falls to $0.157$ at category level, below the $0.6$ width threshold read as an error tolerance. The two objects answer different questions: an aggregated point estimate from a low-bias learner can be precise enough for a decision even though the honest interval built around the same aggregate --- which must also carry the estimated design-related dispersion $\hat\tau$ --- is not. The negative conclusion of this section concerns the interval, not the point estimate; a decision rule based on RMSE alone would read Table~\ref{tab:levels} more favourably, at the cost of not reporting the uncertainty that motivates Sections~\ref{sec:decomposition} and \ref{sec:hierarchical} in the first place.

\section{An application to real scanner data}\label{sec:realdata}

The preceding sections isolate a mechanism entirely within one synthetic generator. This section asks a narrower question that does not require knowing a true elasticity: does a real retail category exhibit a pricing regime and a dispersion pattern of the kind the simulations describe, when the same two learners, the same variation index $V$, the same frontier relation \eqref{eq:scaling} and the same operational threshold are pointed at real transactions instead of simulated ones? The pipeline is implemented as an extension of the same reproducibility codebase used throughout the paper \citep{pricingpanel}. We emphasize at the outset that what follows is illustrative evidence from one category and one retail chain, not an external validation of the paper's simulated magnitudes; \S\ref{sec:reallimits} states the scope restrictions explicitly.

\subsection{Data, pipeline check, and admissibility}

We use the Dominick's Finer Foods database, a weekly store$\times$UPC movement file from a Chicago-area supermarket chain, distributed by the Kilts Center for Marketing at the University of Chicago Booth School of Business \citep{kiltsdominicks}. We analyse the full Soft Drinks category: $8{,}907{,}103$ observations, $83$ stores, $15$ nominal price zones, $1{,}492$ UPCs, and $390$ observed weeks. Before touching these data, the same pipeline is run on synthetic panels built with the exact shape of the real file ($24$ UPCs, $40$ stores, $4$ zones, $250$ weeks, a known injected between-UPC standard deviation of $0.300$) to confirm that it recovers a known truth: \textsf{sieve} returns bias $+0.027$, RMSE $0.082$ and a correlation with the truth of $0.969$, while \textsf{gbr} returns bias $+0.342$, RMSE $0.530$ and a correlation of $0.553$. This reproduces, on a shape-matched synthetic panel rather than on the simulation grid of \S\ref{sec:aggregation}--\S\ref{sec:level}, the same ordering between the two learners, which is why the real-data results below are not read as an artefact of one particular implementation of the estimator.

Applying the paper's pre-estimation identification filter (treatment coefficient of variation above $0.03$ and at least four price movements, as in Table~\ref{tab:frontier}) admits $77.9\%$ of the category's UPCs. Roughly one fifth of the products in this real category do not clear the minimum variation bar the paper already uses to decide whether $\theta$ is worth estimating at all.

\subsection{The observed pricing regime}\label{sec:realregime}

Table~\ref{tab:realregime} reports category-level pricing-regime statistics, defined exactly as in \S\ref{sec:generator} and \S\ref{sec:frontier}.

\begin{table}[t]
\centering
\small
\caption{Category-level pricing-regime statistics for Soft Drinks, defined as in \S\ref{sec:generator} and \S\ref{sec:frontier}. \emph{Effective independent} units follow the effective-sample-size accounting discussed in the Introduction \citep{crump2009dealing,aronowsamii2016does}.}
\label{tab:realregime}
\begin{tabular}{lrl}
\toprule
Metric & Soft Drinks & Simulated reference \\
\midrule
$n_{\text{moves}}$ per $120$-wk window        & $9.98$   & baseline $10$ \\
Mean move magnitude $|\Delta\log P|$          & $23.8\%$ & $3.0$--$7.5\%$ \\
Regime index $V$ (category mean)              & $1.045$  & $\approx8\times10^{-5}$--$4.7\times10^{-2}$ \\
Intra-zone corr.\ of $\Delta\log P$           & $0.603$  & --- \\
Inter-zone corr.\ of $\Delta\log P$           & $0.572$  & $\approx0.01$ \\
Cross-UPC corr.\ of $\Delta\log P$            & $0.011$  & --- \\
Effective independent zones                   & $4.46/15$ ($30\%$) & --- \\
Effective independent UPCs                    & $148.4/1{,}492$ ($10\%$) & --- \\
Pass-through $\hat\rho$ (wholesale$\to$shelf) & $0.457$  & --- \\
\bottomrule
\end{tabular}
\end{table}

The count of regular price changes per $120$-week window, $9.98$, is close to the paper's baseline of ten moves; the mean movement magnitude, $23.8\%$, is several times larger than the $3.0$--$7.5\%$ simulated range, and the resulting index $V=1.045$ lies above the range explored by the simulation grid of \S\ref{sec:frontier} (approximately $V\in[8\times10^{-5},4.7\times10^{-2}]$). We report this gap rather than recalibrate the simulation to match it: it is useful information for designing future simulations, not evidence that the present grid was mis-specified for the purpose it was built for.

The more consequential number in the table is the correlation of $\Delta\log P$ across nominal price zones, $0.572$, nearly as large as the intra-zone correlation of $0.603$; the analogous cross-UPC correlation is much smaller, $0.011$. On the same accounting, the chain's $15$ nominal zones behave as $4.46$ effectively independent zones ($30\%$), and its $1{,}492$ nominal UPCs behave as $148.4$ effectively independent products ($10\%$). This is a direct, real-data instance of the point raised in the Introduction's discussion of effective variation and effective sample size \citep{crump2009dealing,aronowsamii2016does}: the nominal count of zones or products overstates the number of independent design draws by a factor of three to ten in this category, well before any estimator is fit.

The estimated pass-through from wholesale to shelf price is $\hat\rho=0.457$ (se $0.0032$, $n\approx8.76$ million observations): fewer than half of wholesale price movements reach the shelf in this category. Combined with the category-level estimate $\hat\theta\approx-2.10$ of \S\ref{sec:realmeta} below, this would imply a consumer elasticity $\hat\varepsilon_c=\hat\theta/\hat\rho\approx-4.59$ under the convolution of Remark~\ref{rem:split}. We do not report $-4.59$ as a credible demand elasticity; we report it as a real-data illustration of exactly the separation Remark~\ref{rem:split} warns about, namely that $\theta$ identifies $\varepsilon_c\rho$ and not $\varepsilon_c$ on its own.

\subsection{Between-unit dispersion versus sampling noise}\label{sec:realmeta}

We next apply the Paule--Mandel construction of \S\ref{sec:hierarchical} across $J=630$ admissible UPC$\times$window cells of the category (up to three non-overlapping $120$-week windows per admissible UPC over the $390$ observed weeks), for each of the two learners. Table~\ref{tab:realmeta} reports the resulting pooled estimate, its standard error, the between-unit dispersion $\hat\sigma_b$, and the width of three interval constructions: the traditional within-panel bootstrap, the hierarchical Paule--Mandel interval of \eqref{eq:pm}, and a posterior-predictive interval that additionally propagates between-unit heterogeneity into a prediction for a new unit.

\begin{table}[t]
\centering
\small
\caption{Paule--Mandel construction of \S\ref{sec:hierarchical} applied across $J=630$ admissible UPC$\times$window cells of the Soft Drinks category. \emph{Width} is the width of the $95\%$ interval under each construction; \emph{ratio} is the hierarchical width divided by the bootstrap width. The within-unit variance component is numerically degenerate at this level of aggregation (\S\ref{sec:reallimits}) and is omitted.}
\label{tab:realmeta}
\begin{tabular}{lrrrrrrrr}
\toprule
Learner & $J$ & $\hat\theta_{\text{PM}}$ & se$_{\text{PM}}$ & $\hat\sigma_b$
        & Boot.\ width & Hier.\ width & Post.\ pred.\ width & Ratio \\
\midrule
\textsf{gbr}   & $630$ & $-2.095$ & $0.054$ & $1.215$ & $0.039$ & $0.213$ & $4.776$ & $5.5\times$ \\
\textsf{sieve} & $630$ & $-2.096$ & $0.097$ & $2.339$ & $0.030$ & $0.380$ & $9.194$ & $12.8\times$ \\
\bottomrule
\end{tabular}
\end{table}

At the category level, $\hat\sigma_b$ is $1.215$ for \textsf{gbr} and $2.339$ for \textsf{sieve}. The corresponding within-unit variance component is numerically degenerate at this aggregate ($\sigma_w^2$ on the order of $10^{14}$--$10^{15}$), so neither its raw value nor the derived ratio $\sigma_b^2/(\sigma_w^2+\sigma_b^2)$ is reported here; the finding below does not depend on that quantity. The defensible comparison is interval width: the hierarchical interval is $5.5$ times wider than the bootstrap for \textsf{gbr} and $12.8$ times wider for \textsf{sieve}, and the posterior-predictive interval is wider still. This is the real-data counterpart of Table~\ref{tab:decomp} and \S\ref{sec:hierarchical}: a within-panel bootstrap constructed from the realised category, without reference to dispersion across units, understates the interval that a working model of between-unit design dispersion would produce, by a wider margin here than in the simulated grid.

\subsection{Confrontation with the simulated frontier}\label{sec:realfrontier}

\begin{figure}[t]
\centering
\includegraphics[width=0.85\textwidth]{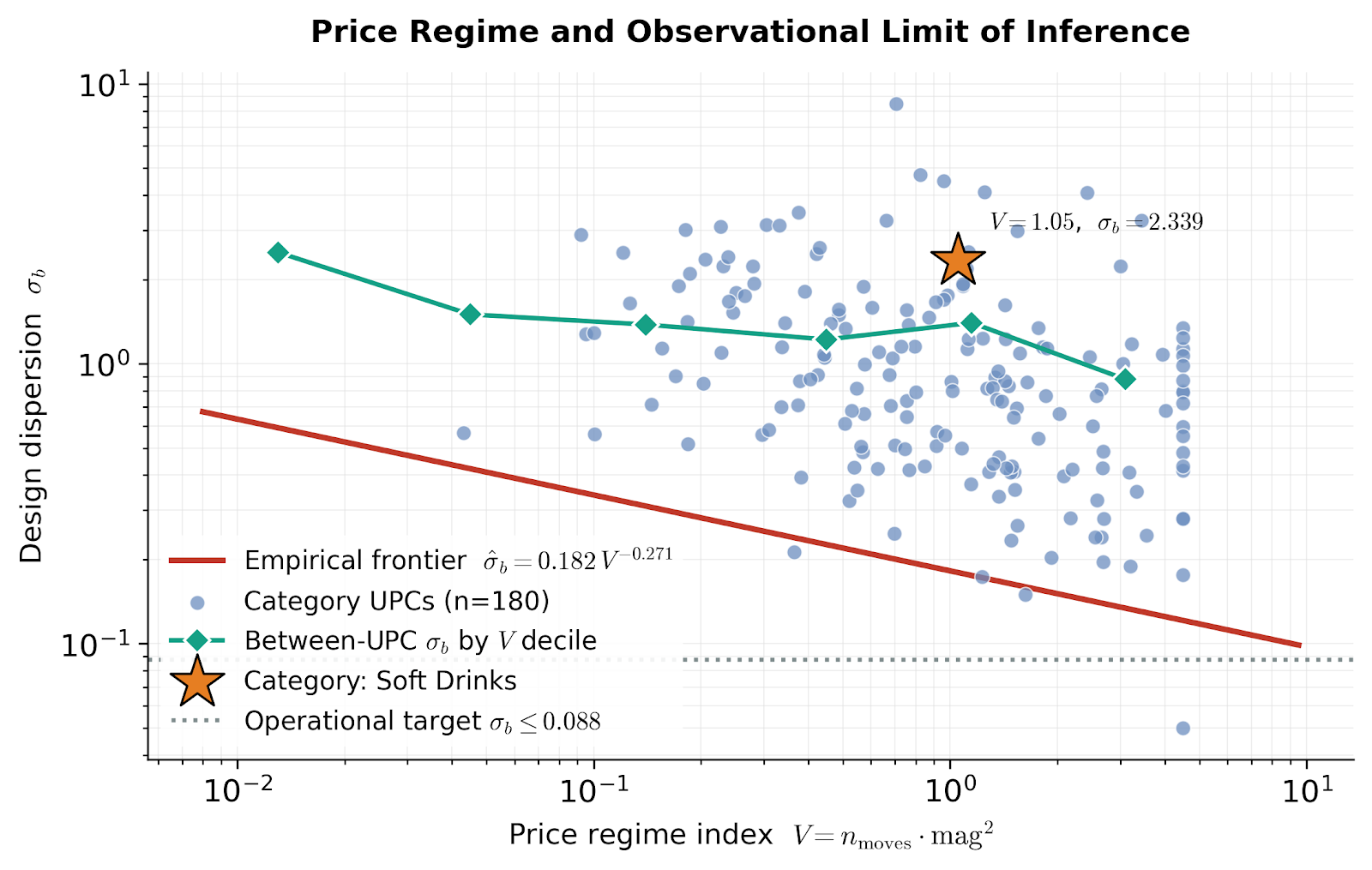}
\caption{Design dispersion $\sigma_b$ against the regime index $V$ for the $180$ Soft Drinks UPCs with enough non-overlapping $120$-week windows to compute an individual $\sigma_b$ (light points), the between-UPC $\sigma_b$ by $V$ decile (green diamonds), and the category aggregate using all $1{,}492$ UPCs and $8.9$ million observations (star). The solid line is the frontier \eqref{eq:scaling} fitted on the \emph{simulated} grid of \S\ref{sec:frontier} and reproduced here without refitting; the dotted line is the paper's $\sigma_b\le0.088$ operational target of \S\ref{sec:implications}. The two lines summarise different sources of evidence plotted on the same axes, not a fit to these data.}
\label{fig:realfrontier}
\end{figure}

Figure~\ref{fig:realfrontier} plots $\sigma_b$ against $V$ for the $180$ UPCs of the category for which enough separate $120$-week windows exist to compute an individual $\sigma_b$ (most UPCs span only two or three non-overlapping windows over $390$ weeks; the estimation of \S\ref{sec:realregime}--\S\ref{sec:realmeta} still uses all $1{,}492$ UPCs and the full $8.9$ million observations, and it is that category aggregate, not a restricted sample, that the star marks). The solid line is the frontier \eqref{eq:scaling} fitted on the simulated grid of \S\ref{sec:frontier}, reproduced here without refitting.

At the category's mean $V=1.045$, the simulated frontier predicts $\hat\sigma_b\approx0.180$; the observed value for \textsf{sieve} at category level is $2.339$, a factor of $13.0$--about an order of magnitude--above the prediction. Grouping the $180$ plotted UPCs into deciles of $V$, between-UPC $\sigma_b$ declines from $2.63$ in the lowest decile to $0.87$ in the highest: the frontier's qualitative direction, more identifying price variation associated with lower dispersion, reproduces in the real category; its level does not. We read the size of the gap as informative rather than as a defect of \eqref{eq:scaling}: \S\ref{sec:frontier} already describes that relation as a regularity of one simulated grid, and \S\ref{sec:realregime} has already identified a mechanism a purely independent-design frontier cannot capture, namely that real zones and real UPCs are considerably more correlated with one another than the independently drawn designs of the simulation. The real $\sigma_b$ therefore mixes design dispersion with cross-unit covariance and with genuine heterogeneity in unit-level elasticities, exactly the failure mode of the exchangeability assumption anticipated in the Introduction's discussion of variance components across units.

\subsection{Spatial versus product aggregation}\label{sec:realagg}

\begin{figure}[t]
\centering
\includegraphics[width=\textwidth]{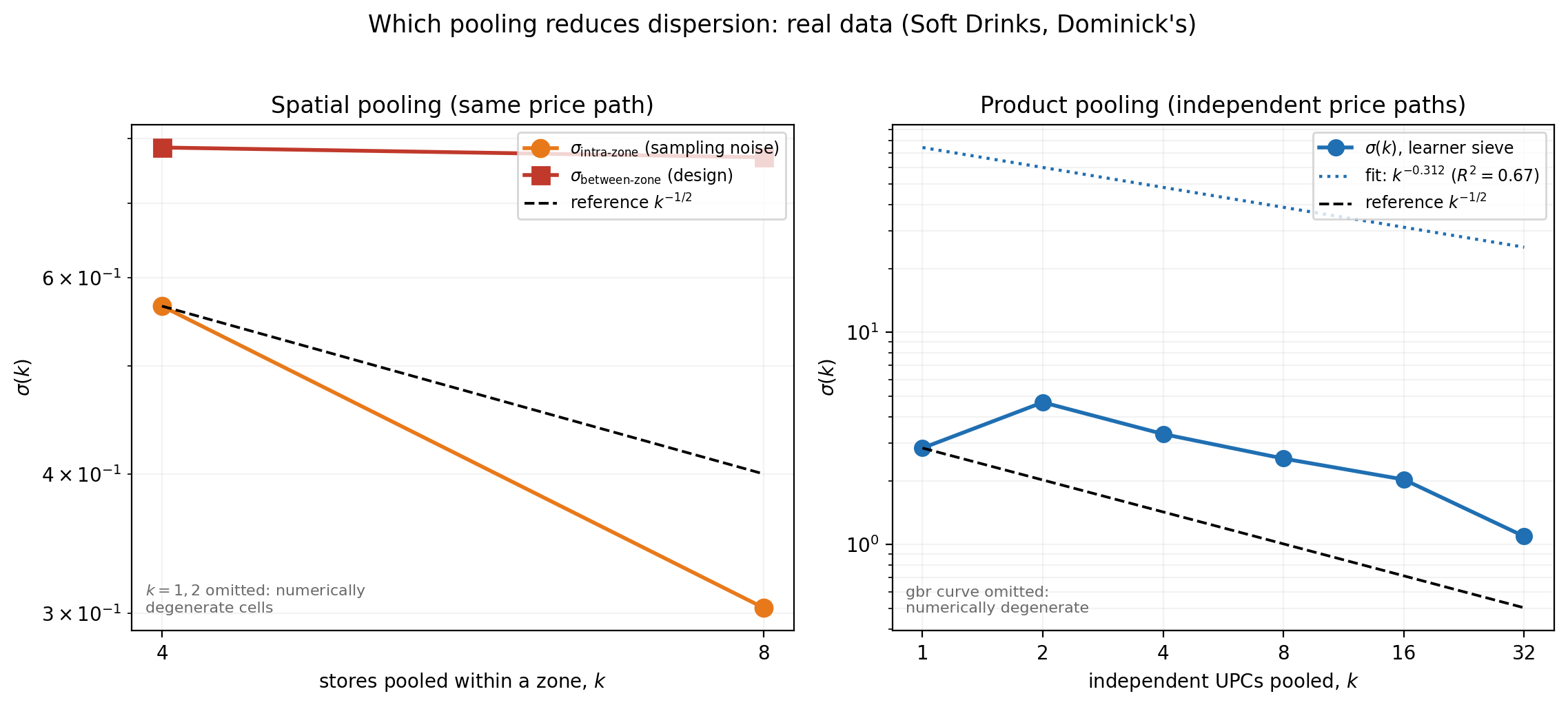}
\caption{Left: within-zone sampling noise $\sigma_{\text{intra-zone}}$ and between-zone design dispersion $\sigma_{\text{between-zone}}$ as stores are pooled within a nominal price zone ($k=4,8$; $k=1,2$ omitted as numerically degenerate). Right: dispersion $\sigma(k)$ for \textsf{sieve} as independently priced UPCs are pooled ($k=1,\ldots,32$; \textsf{gbr} omitted as numerically degenerate), with the fitted log--log slope and the $k^{-1/2}$ independent-design reference of Proposition~\ref{prop:agg}.}
\label{fig:realagg}
\end{figure}

Figure~\ref{fig:realagg} repeats the aggregation exercise of \S\ref{sec:aggregation} and Proposition~\ref{prop:agg} on the real category, separating stores pooled within the same nominal zone (a common price path) from independently priced UPCs.

Pooling stores within a zone leaves the between-zone design dispersion essentially flat, $0.785$ at $k=4$ stores to $0.769$ at $k=8$, while the within-zone sampling noise falls from $0.566$ to $0.304$ as $k$ doubles, in the direction expected from pooling within-zone sampling noise. This is the store-level, real-data analogue of Proposition~\ref{prop:agg} under near-perfect correlation: pooling units that share one realised price path reduces outcome noise without creating additional identifying variation. Pooling independent UPCs instead, and restricting to \textsf{sieve} because \textsf{gbr}'s product-level aggregation curve is numerically degenerate here and is not reported, gives a fitted log--log slope of $-0.312$ ($R^2=0.67$) over $k=1,2,4,8,16,32$ UPCs--the raw points are not perfectly monotonic, $\sigma(k)$ rises from $k=1$ to $k=2$ before declining through $k=32$, consistent with a moderate fit rather than an exact power law--weaker than the $-1/2$ independent-design reference and weaker than the slope \textsf{sieve} attains on the simulated grid. The most defensible reading is that price trajectories of different UPCs in this category are more correlated with one another, through shared promotional calendars and category-wide campaigns, than the approximately independent product-level designs assumed in \S\ref{sec:aggregation}; the direction of the benefit from pooling independent products is preserved, but its rate is slower than $k^{-1/2}$.

\subsection{Operational classification and the cost of redesign}\label{sec:realclass}

\S\ref{sec:eight} pre-specifies an operational width threshold of $W\le0.6$ for a decision-grade interval; \S\ref{sec:hierarchical}--\S\ref{sec:level} distinguish a within-panel bootstrap, which does not represent between-unit design dispersion, from an ``honest'' interval that does. Crossing these two distinctions gives three outcomes for any estimated unit: \emph{invalid-narrow}, where the bootstrap alone falls under the threshold although \S\ref{sec:decomposition} and \S\ref{sec:hierarchical} indicate it should not be read as capturing between-unit dispersion; \emph{honest-inconclusive}, where the hierarchical interval represents that dispersion but exceeds the threshold; and \emph{honest-decisive}, where the hierarchical interval both represents that dispersion and clears the threshold. Table~\ref{tab:realclass} applies this classification to the category at three levels of aggregation: individual UPC, brand (proxied by the UPC manufacturer prefix, not a verified brand table), and the category as a whole.

\begin{table}[t]
\centering
\small
\caption{Operational classification of \eqref{eq:pm}-based intervals against the $W\le0.6$ threshold of \S\ref{sec:eight}, at three levels of aggregation. \emph{Brand} is proxied by UPC manufacturer prefix ($40$ brands). \emph{Invalid-narrow}: bootstrap interval under threshold. \emph{Honest-inconclusive}: hierarchical interval over threshold. \emph{Honest-decisive}: hierarchical interval under threshold.}
\label{tab:realclass}
\begin{tabular}{llrrrr}
\toprule
Level & Learner & Invalid-narrow & Honest-inconclusive & Honest-decisive & Total \\
\midrule
UPC      & \textsf{gbr}   & $50$ ($7.9\%$)  & $564$ ($89.5\%$) & $16$ ($2.5\%$) & $630$ \\
UPC      & \textsf{sieve} & $85$ ($13.5\%$) & $507$ ($80.5\%$) & $38$ ($6.0\%$) & $630$ \\
\addlinespace
Brand    & \textsf{gbr}   & $20$ ($50.0\%$) & $15$ ($37.5\%$)  & $5$ ($12.5\%$) & $40$ \\
Brand    & \textsf{sieve} & $23$ ($57.5\%$) & $14$ ($35.0\%$)  & $3$ ($7.5\%$)  & $40$ \\
\addlinespace
Category & both           & $0$             & $0$              & $1$ ($100\%$)  & $1$ \\
\bottomrule
\end{tabular}
\end{table}

\begin{figure}[t]
\centering
\includegraphics[width=0.85\textwidth]{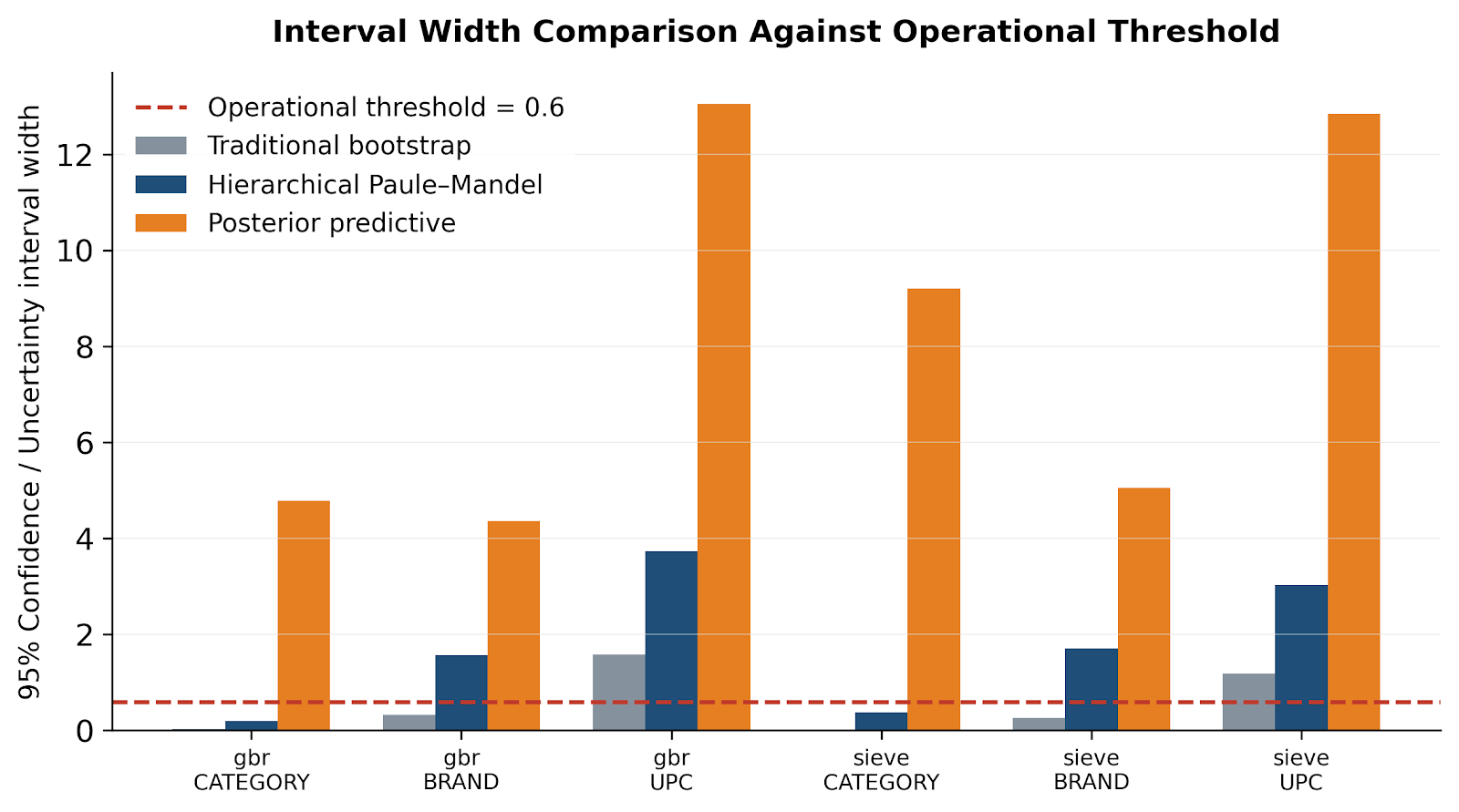}
\caption{Width of the $95\%$ interval under three constructions--traditional within-panel bootstrap, hierarchical Paule--Mandel, and posterior predictive--for both learners at the UPC, brand, and category levels, against the $W=0.6$ operational threshold of \S\ref{sec:eight}.}
\label{fig:realintervals}
\end{figure}

Figure~\ref{fig:realintervals} shows the underlying interval widths against the $0.6$ threshold for both learners at all three levels.

At UPC level, between $80\%$ and $90\%$ of products are honest-inconclusive and between $8\%$ and $14\%$ are invalid-narrow; only $2.5\%$ to $6\%$ are honest-decisive. At brand level roughly half of the $40$ brands remain invalid-narrow. Only at the level of the category as a whole, where all admissible UPCs are pooled into one estimate, do both learners reach honest-decisive. This is the real-data counterpart of Tables~\ref{tab:hier} and \ref{tab:nivel}: in this category, the aggregation level at which an honest interval is also decisive is far coarser than the individual product.

Applying Proposition~\ref{prop:agg}'s $\sqrt{k}$ logic in reverse, as in the illustrative calculation of \S\ref{sec:implications}, the fifteen brands presently classified honest-inconclusive would need approximately $21$ to $22$ independently priced zones to bring their interval under the $0.088$ target, comparable to, though somewhat more than, the fifteen zones this chain already operates. At category level the same target requires approximately $(2.339/0.088)^2\approx707$ independent zones for \textsf{sieve}, forty-seven times the fifteen the chain has. The gap between these two numbers is itself informative: closing it at brand level is a plausible design change for this chain, while closing it at category level, under this working model and this level of between-unit correlation, is not.

\subsection{Scope of this illustration}\label{sec:reallimits}

The evidence in this section comes from one category (Soft Drinks) in one retail chain (Dominick's Finer Foods, Chicago, 1990s) and should not be extrapolated to retail categories, chains, or time periods in general. Unlike the simulations, there is no known $\theta$ against which to measure coverage directly; the notebook behind this section instead reports a holdout-based proxy, contrasting an interval built from $k$ UPCs against the mean of excluded UPCs, which is an honest diagnostic under that specific construction and not nominal coverage against a ground truth. The category-level within-unit variance component, the associated ratio $\sigma_b^2/(\sigma_w^2+\sigma_b^2)$, and \textsf{gbr}'s product-aggregation slope are numerically degenerate, as noted above, and are not treated as substantive estimates anywhere in this section. ``Brand'' is approximated from the UPC's manufacturer prefix rather than from a verified brand table, and the pass-through and promotional-window calculations of \S\ref{sec:realregime} rely on quantity and profit fields that are not present in every extract of this database. With these qualifications, the pattern is consistent with the paper's central claim rather than a demonstration of it: the direction of every comparison in \S\ref{sec:realregime}--\S\ref{sec:realclass} matches the simulated mechanism, while several of the levels do not, for reasons the mechanism itself anticipates.

\section{Implications for data design}\label{sec:implications}

\paragraph{Adding rows is not equivalent to adding independent treatment variation.} When all regions face the same national list-price path, additional regions can still reduce outcome noise, help estimate nuisance functions, and improve precision conditional on that path. What they do not provide is an additional independent realisation of the price trajectory. Proposition~\ref{prop:agg} makes the relevant distinction explicit: the gain from aggregation is governed by the covariance of design-specific errors, not by the raw number of rows.

\paragraph{Longer panels help only through the variation they actually add.} In the simulation grid, larger $V=n_{\text{moves}}\times\text{magnitude}^2$ is associated with lower across-design dispersion. If the fitted relation \eqref{eq:scaling} were used descriptively, doubling $V$ would be associated with a reduction in $\sigma_b$ by a factor of about $2^{0.271}=1.21$. Because the exponent is a simulation regularity and $V$ is only a heuristic index, this arithmetic should not be used as a general forecast for extending a real panel.

\paragraph{Independent pricing paths can be more valuable than replicated exposure to one path.} In the simulated portfolio, product units with separately generated price trajectories display an approximately $\sqrt{k}$ reduction in design-specific dispersion. The key empirical requirement is low covariance of the relevant design errors; distinct product labels alone do not guarantee it. In an application, the covariance of price-setting rules, common cost shocks and synchronized promotions would need to be assessed before treating product histories as independent design draws.

\paragraph{Randomisation is one transparent way to create and justify independence.} Controlled regional price assignment can generate treatment variation whose assignment mechanism is known, thereby supporting design-based reasoning in a way that passive histories generally cannot. Under the illustrative assumptions of Proposition~\ref{prop:agg}, taking the baseline $\sigma_b=0.4815$ and imposing independent, mean-zero regional design errors would imply $0.4815/\sqrt{30}\approx0.088$ for an average across thirty regions. This is an algebraic illustration, not a simulated result for a thirty-region experiment. Natural experiments, staggered policy changes, or other sources of plausibly exogenous and weakly dependent price variation could play a similar identifying role. The central recommendation is therefore to create or exploit independent identifying variation, not that randomisation is the only admissible design.

\paragraph{What this arithmetic costs in a real category.} \S\ref{sec:realclass} applies the same $\sqrt{k}$ logic to a real category rather than to the baseline simulation, using the same $0.088$ target. At brand level, closing the gap requires on the order of $21$--$22$ independently priced zones, close to the fifteen this chain already operates; for the category as a whole it requires on the order of $707$, forty-seven times as many. The two figures are not predictions for any other chain or category, but they illustrate, with real numbers, how sharply the cost of independent identifying variation can depend on the level at which the estimate is ultimately reported.

\section{Limitations}\label{sec:limitations}

\paragraph{Synthetic evidence.} Every quantitative result is measured on a generator written for this study. This is useful because repeated price trajectories and design-specific ground truth are observable by construction, but it sharply limits external validity. The simulations establish that the proposed mechanism can be quantitatively important under the stated DGP; they do not establish that between-design dispersion dominates for every short pricing panel or every estimator.

\paragraph{The error decomposition must match the implemented ground truth.} The formal decomposition in Section~\ref{sec:decomposition} is defined for $e(D,u)=\hat\theta(D,u)-\theta(D)$. This matters because the generator allows the design-specific truth to vary through pass-through. Any implementation that computes $\sigma_b$ from $\operatorname{Var}_D\{\mathbb{E}_u[\hat\theta\mid D]\}$ without subtracting $\theta(D)$ would combine variation in the estimand with variation in estimation error. Reproducibility code should therefore report explicitly which object is used in the decomposition.

\paragraph{A heuristic variation index, not an information measure.} The index $V=n_{\text{moves}}\times\text{magnitude}^2$ is a useful low-dimensional summary of the simulation grid, but the paper does not derive it as Fisher information or prove that estimator dispersion must scale as a power of $V$. Consequently, the fitted exponent $-0.271$ is a descriptive simulation coefficient and extrapolations from \eqref{eq:scaling} are illustrative only.

\paragraph{Variance-component assumptions.} Interpreting Paule--Mandel's $\tau^2$ as design variance requires approximately independent or weakly dependent design draws, adequate within-unit variance estimates, exchangeability of the design-specific displacement and, for a clean interpretation, common unit-level truths. When true elasticities vary, $\hat\tau^2$ generally mixes genuine heterogeneity with design uncertainty. With only eight units in the main homogeneous exercise, uncertainty in $\hat\tau^2$ itself can also be material.

\paragraph{One estimator implementation.} The simulations use the estimator in \eqref{eq:dml}, including median aggregation of fold-specific residual-on-residual coefficients. This is DML-motivated but is not the canonical pooled DML2 estimator. The canonical DML theory also relies on regularity and dependence conditions that are not established here for the short serially dependent panel. The paper should therefore avoid attributing the simulation findings to Double Machine Learning as a class. Whether analogous between-design dispersion arises for instrumental-variable, structural, Bayesian, or alternative orthogonal-score estimators remains an empirical and theoretical question.

\paragraph{Calibration and decision thresholds.} The $W\le0.6$ precision threshold is a pre-specified operational criterion for the simulation exercise, not a universal standard for pricing decisions. Likewise, the identification filter is calibrated on the same generator and has limited transportability across configurations. We quantify this directly: freezing the filter's thresholds at the values chosen under the calibration configuration and applying them unmodified to three other configurations, precision --- the share of admitted panels whose point estimate is within $0.25$ of the truth --- falls from $0.39$--$0.42$ in the calibration configuration to $0$ (no panel is admitted, or none of the few admitted is useful) under a low move-magnitude configuration, and to $0.15$--$0.29$ under a low move-count configuration. These thresholds are useful for organizing the experiment but should not be interpreted as externally validated cutoffs, and the drop in precision away from the calibration configuration is itself evidence of that limited transportability rather than only a caveat about it.

\begin{table}[t]
\centering
\caption{Recovery of the injected elasticity on the independently coded legacy generator described in Appendix~\ref{app:dgp} (\emph{the second generator}), five hundred replications per regime. $\theta$ true is $\beta\rho$ as realised by that generator. The \emph{round point} regime pins the shelf price so that $\rho=0$ and $\theta=0$ by construction; the large median estimate and RMSE reported for it illustrate the diagnostic failure this configuration is designed to produce, not a usable point estimate.}
\label{tab:recovery}
\begin{tabular}{lrrrrrr}
\toprule
Regime & $\beta$ & $\rho$ & $\theta$ true & $\hat\theta$ median & Bias & RMSE \\
\midrule
clean       & $-1.30$ & $0.85$ & $-1.105$ & $-0.980$ & $0.125$  & $0.222$ \\
weak        & $-1.00$ & $0.85$ & $-0.850$ & $-0.149$ & $0.701$  & $0.824$ \\
confounded  & $-1.40$ & $0.85$ & $-1.190$ & $-0.115$ & $1.075$  & $1.172$ \\
collinear   & $-1.20$ & $0.85$ & $-1.020$ & $\phantom{-}0.483$ & $1.503$ & $1.952$ \\
round point & $-1.10$ & $0.00$ & $\phantom{-}0.000$ & $-4.688$ & $-4.688$ & $6.089$ \\
\bottomrule
\end{tabular}
\end{table}

Table~\ref{tab:recovery} shows that recovery on this independently coded generator is reasonable in the \textsf{clean} regime --- the one every other table in this paper uses --- and degrades sharply once confounding, collinearity, or a pinned shelf price are introduced, none of which are surprising on their own but which bound how far the paper's quantitative claims should be expected to travel outside the \textsf{clean} configuration.

\paragraph{External validation.} As a check that falls short of real data but is stronger than relying on one generator, we compared the recovery of the injected elasticity in the \textsf{clean} regime across the two independently written data-generating processes: the extended generator used throughout this paper and the legacy generator behind Table~\ref{tab:recovery}. The two agree closely (bias difference $0.077$ against a pre-specified tolerance of $0.10$), which is evidence that the paper's central mechanism is not an artifact of one specific implementation, though it remains evidence from synthetic data on both sides. Section~\ref{sec:realdata} takes the further step of computing the same variation summaries and dependence diagnostics on a real price history---a full retail category from the Dominick's Finer Foods database---rather than on a second synthetic generator. The qualitative pattern of the mechanism studied in this paper reproduces there; several of its quantitative magnitudes do not, for reasons that section discusses in detail, and the exercise remains evidence from one category and one chain rather than a general external validation.

\section{Conclusion}\label{sec:conclusion}

This paper uses a synthetic pricing environment to separate two sources of inferential uncertainty that are easy to conflate in short panels. The first is variation conditional on one realised price trajectory. The second is variation in the estimator's conditional mean error across alternative trajectories generated by the pricing process. In the baseline DGP, the second component is quantitatively large: for the gradient-boosted specification it accounts for $97.6\%$ of the variance of the estimation error. The number is simulation-specific, but the decomposition clarifies why an interval based on one realised panel need not represent repeated-design performance.

The simulation evidence also refines the diagnosis of undercoverage. The moving-block bootstrap standard error is not simply too small relative to within-design dispersion; in fact it exceeds the measured $\sigma_w$ in the baseline decomposition. The more important feature is that the estimator's conditional centre varies across realised price histories. None of the eight within-panel constructions evaluated here reaches nominal coverage, although this finite comparison is not an impossibility theorem. The result motivates explicitly modelling or sampling the distribution of designs rather than treating a wider conventional standard error as a complete solution.

Two constructive findings follow. First, the gain from aggregation depends on the covariance of design-specific errors. Proposition~\ref{prop:agg} shows exactly why independent design draws produce a $\sqrt{k}$ reduction in standard deviation and why a common component limits that gain. The simulated product units are close to the independent benchmark, whereas additional exposure to a common price path does not create the same source of identifying variation. This is the precise sense in which adding observations is not equivalent to adding identification.

Second, a standard between-unit variance component can be useful when multiple independently realised price paths are available. In the homogeneous simulation, the Paule--Mandel-based variance augmentation raises empirical coverage from $0.469$ to $0.931$ for the posterior-centred construction, at the cost of much wider intervals. This exercise does not establish a new meta-analytic estimator. It shows that, under a random-bias working model, information across designs can represent a component that a single realised trajectory cannot identify nonparametrically. When true unit effects differ, the between-unit component mixes heterogeneity with design uncertainty and must be interpreted more cautiously.

The fitted relation $\hat\sigma_b\approx0.182V^{-0.271}$ provides a compact summary of the simulation grid, not a scaling law. Its role is descriptive: within the explored DGP, substantially more independent price movement is associated with lower across-design dispersion, but the decline is slow over the observed range. The broader empirical lesson is therefore about data design. If the inferential target requires performance across plausible future price histories, collecting more rows under essentially the same treatment path may be less valuable than generating additional, credibly independent price variation. Controlled regional randomisation is one transparent way to create such variation; other quasi-experimental sources may serve the same purpose when their assignment and dependence structure are defensible.

The paper's contribution is consequently best read as a shift in emphasis from ``which standard error should be attached to this one passive panel?'' toward ``what variation would make the elasticity identifiable and its uncertainty learnable at the decision level?'' The simulations suggest that, in sparse pricing regimes, better inference and better data design are closely linked. Section~\ref{sec:realdata} takes a first step toward establishing how far that conclusion travels beyond the present generator; extending it to other categories and retail settings remains the next empirical task.

\paragraph{What the real category adds.} The application in Section~\ref{sec:realdata} is not a second simulation and not a validation exercise: applied to a full Dominick's Finer Foods category, the same pipeline finds nominal price zones and nominal products far more correlated with one another than the independent designs assumed in the simulated grid, and an honest interval wider, relative to a conventional bootstrap, than anywhere in that grid. Holding this real illustration alongside the simulation is what lets the paper say, with the appropriate qualification, that the mechanism documented here is not solely an artefact of the generator that supplies most of its numbers---while continuing to rely on that generator, and not on the real category, for every quantitative claim about magnitudes, rates, and coverage.

\paragraph{Reproducibility.} The generator, the estimators, the thirty-five experiments and the pre-registered decision rules that produced every figure in this paper are published as an executable notebook. All results are obtained from synthetic data and no commercial information is used.

\clearpage

\appendix

\section{The data-generating process}\label{app:dgp}

The data generating process (DGP; \cite{pricingpanel}) produces a weekly panel at the unit $\times$ region $\times$ week
level. Volume is drawn against the shelf price, which is generated from the list
price at a pass-through rate that may be constant, heterogeneous across regions,
or dependent on the shopkeeper's margin state. Prices, promotions, competitor
prices, costs, weather, holidays and an inflation index are generated jointly so
that confounding is present by construction. The demand shock is drawn from a
dedicated random stream, which is what allows a design to be held fixed while
the shock is resampled (Section~\ref{sec:decomposition}). All monetary
quantities are expressed relative to a reference list price $p_0$; the panel
starts on 2 January 2023 and runs $W$ Mondays.

\subsection{Exogenous series}

For week $t = 0,\dots,W-1$ the inflation index is $I_t = 100\,(1.0016)^t(1+\nu_t)$
with $\nu_t \sim N(0, 0.002^2)$; holidays per month and working days
($22$ minus holidays) come from a fixed calendar map; national real income is
$S^{\text{nac}}_t = 100\,(1.0015)^t$ and its regional counterpart is
$S_{rt} = S^{\text{nac}}_t \exp(\sum_{s\le t} \zeta_{rs})$ with
$\zeta_{rs}\sim N(0,\sigma_S^2)$, which is what gives the cash-pressure index
a regional component. Weather is $\text{clima}_{rt} = T_r + 6\sin(2\pi\,
\text{woy}_t/52 - 1.2) + N(0,1)$, with $T_r \sim U(13,31)$ drawn once per
region. Variable cost starts at $0.50p_0$ and
receives four step shocks of $\pm(2\text{--}6)\%\,p_0$ at random weeks, clipped
to $[0.30, 0.62]\,p_0$.

\subsection{Prices, promotions and the competitor}

Promotions are $6$ to $10$ windows of $2$--$3$ weeks per series; the flag is
$\text{promo}_{rt}\in\{0,1\}$ and the intensity is
$\text{pint}_{rt} = \pi_{ir}\,\text{promo}_{rt}$, where $\pi_{ir} \sim U(0.10,
0.25)$ is drawn once per series, so a promoted week is discounted by the same
amount throughout a series. The list price is a step function,
\begin{equation}
P^{\text{si}}_{irt} \;=\; \operatorname{clip}\!\Big(p_0 + \textstyle\sum_{k \le t}
\delta_{irk},\; 0.72p_0,\; 1.55p_0\Big),
\end{equation}
where $n_{\text{moves}}$ weeks are drawn without replacement from
$[6, W-4]$ and each contributes a jump $\delta = \pm U(\text{mag})\,p_0$ with a
random sign. Two switches control the confounding of the treatment. With
$\text{promo\_conf} = 0$ a move that falls on a promotion week is discarded, so
price and promotion are orthogonal by construction; with $\text{promo\_conf} > 0$
the move is relocated \emph{onto} a promotion week with that probability, which
is how the confounded regime is produced. In the collinear regime the list price
is replaced by $p_0 (C_{rt}/1.05p_0)^{0.95}$, a near-deterministic function of
the competitor's price.

The competitor's price is a random walk around $1.05p_0$ with weekly standard
deviation $0.004p_0$, plus $\text{comp\_moves}$ discrete jumps of
$\pm U(4\%,9\%)\,p_0$; when $\kappa = \text{comp\_react} > 0$ it also follows the
own price with one week's lag,
$C \leftarrow C\exp\!\big(\sum_{s\le t}\kappa\,\Delta \log P^{\text{si}}_{s-1}\big)$,
and the result is clipped to $[0.70, 1.45]\,p_0$. Without discrete jumps the
competitor's log price has a standard deviation of about $1.5\%$ over $120$
weeks, below the demand shock, and the cross-price channel is not identified by
construction; this is why $\text{comp\_moves} = 8$ in every experiment run on this
generator.

\subsection{Pass-through and the shelf price}

The shopkeeper's price is
\begin{equation}
P^{\text{so}}_{irt} \;=\; 1.12\,p_0
\left(\frac{P^{\text{si}}_{irt}}{p_0}\right)^{\rho_{irt}}
\big(1 - 0.6\,\text{pint}_{rt}\big),
\label{eq:pt}
\end{equation}
with a pass-through that is heterogeneous across regions and dependent on the
retailer's margin state,
\begin{equation}
\rho_{irt} \;=\; \operatorname{clip}\!\left(\bar\rho_{ir}
\left[1 + \lambda\,\frac{m^\ast - M_{ir,t-1}}{m^\ast}\right],\, 0,\, 1.2\right),
\qquad
\bar\rho_{ir} = \operatorname{clip}(\rho_i m_r,\,0,\,1),
\label{eq:rho}
\end{equation}
where $M_{ir,t-1} = (P^{\text{so}}_{ir,t-1} - P^{\text{si}}_{irt}) /
P^{\text{so}}_{ir,t-1}$ is the retailer's margin carried into the week,
$m^\ast = 0.12$ is the target margin, $m_r = \operatorname{clip}(1 +
N(0,\sigma_\rho^2), 0.25, 1.6)$ is drawn once per region, and $\lambda$ governs
the state dependence: when the margin falls below target the shopkeeper
transmits more. In the round-point
regime the shelf price is pinned, $P^{\text{so}} = 1.12p_0(1-0.6\,\text{pint})$,
so $\rho = 0$ and the list price does not enter demand at all.

Because \eqref{eq:rho} makes $\rho$ a function of the realised series,
$\rho_i$ is not the estimand. The generator therefore computes the
\emph{realised} pass-through of each series,
\begin{equation}
\rho^{\text{eff}}_{ir} \;=\;
\frac{\operatorname{Cov}\!\big(\log \tilde P^{\text{so}}_{irt},\,
\log (P^{\text{si}}_{irt}/p_0)\big)}
{\operatorname{Var}\!\big(\log (P^{\text{si}}_{irt}/p_0)\big)},
\qquad
\tilde P^{\text{so}} = \frac{P^{\text{so}}}{1.12p_0(1-0.6\,\text{pint})},
\end{equation}
and stores it in the panel. The ground truth against which every table in this
paper measures is $\theta_i = \beta_i \,\bar\rho^{\text{eff}}_i$, read from the
generated panel rather than from its configuration; comparing against $\beta_i$
or against the configured $\rho_i$ would report generator dispersion as
estimator bias.

\subsection{Demand}

Sell-out volume is
\begin{equation}
\begin{split}
\log Q^{\text{so}}_{irt} \;=\; \log 1000
&+ \beta_{it} \log\frac{P^{\text{so}}_{irt}}{1.12 p_0}
 + 0.30 \log\frac{C_{rt}}{1.05 p_0}
 + 0.40\,\text{promo}_{rt} \\
&+ 0.05\,(\text{clima}_{rt} - T_r)
 + 0.10 \log\frac{I_t}{100}
 + 0.05\,\text{cp}_{rt}
 + u_{irt},
\end{split}
\label{eq:demanda}
\end{equation}
with $u_{irt} \sim N(0,\sigma_u^2)$ and $\text{cp}_{rt}$ a cash-pressure index
built as a fortnightly calendar term minus standardised log regional income. The
elasticity profile $\beta_{it}$ is constant unless a structural break
$(t^\ast, \beta^{\text{post}})$ is injected. When an asymmetry $a$ is requested
the level term is replaced by a Houck decomposition in the first differences of
$\log(P^{\text{so}}/1.12p_0)$: $\beta(1-a)\sum_{s\le t}(\Delta \log
P^{\text{so}}_s)^+ + \beta(1+a)\sum_{s\le t}(\Delta \log P^{\text{so}}_s)^-$.

Sell-in volume adds forward buying and, optionally, order refusal:
$Q^{\text{si}} = Q^{\text{so}}(1 + f_{irt}) + N(0,0.02^2)\,Q^{\text{so}}$, where
$f = 0.30$ in the two weeks preceding each list-price rise above $2\%$. Note
that \eqref{eq:pt} and \eqref{eq:demanda} together imply
$\partial \log Q / \partial \log P^{\text{si}} = \beta\rho$, which is the
convolution discussed in Remark~\ref{rem:split}.

\paragraph{Nominal and deflated prices.} The panel stores prices in nominal
terms, multiplied by $I_t/100$, while demand \eqref{eq:demanda} is generated
against the deflated series. The estimator's treatment is therefore the nominal
log list price, and log inflation is among the controls $W$, so the deflation
cancels in the partialled-out regression; we record the asymmetry because it is
the kind of detail that decides whether a replication reproduces a bias of
$0.15$ or of $0.20$.

\paragraph{Seed separation.} Two independent generators of randomness are used.
The \emph{design} stream draws prices, promotions, the competitor, costs,
weather, income and the regional pass-through multipliers; the \emph{shock}
stream draws $u_{irt}$ and the sell-in noise. Fixing the first and varying the
second resamples the demand shock while holding the design fixed, which is the
mechanism of Section~\ref{sec:decomposition}. All other experiments vary both.

\subsection{Regimes, portfolio and baseline parameters}

A regime is a configuration of the price-generating block:
\textsf{clean} ($n_{\text{moves}} = 10$, magnitude $3.0$--$7.5\%$, no
confounding), \textsf{weak} ($2$ moves of $0.4$--$0.9\%$), \textsf{confounded}
(as clean but $\text{promo\_conf} = 0.9$), \textsf{collinear} (list price
following the competitor at $0.95$), and \textsf{round point} (as clean but with
the shelf price pinned, so $\theta = 0$). The default portfolio has eight units
in four brands and two categories, with consumer elasticities from $-0.90$ to
$-1.40$ and configured pass-through from $0$ to $0.85$, spanning the five
regimes. Experiments that need a controlled topology substitute their own
portfolio: Section~\ref{sec:aggregation} uses eight \textsf{clean} units in two
brands of one category ($\beta = -1.30$ and $-1.00$), Section~\ref{sec:hierarchical}
eight \textsf{clean} units sharing one brand, and Section~\ref{sec:level} six
categories $\times$ two brands $\times$ two units, all \textsf{clean}.

\begin{table}[t]
\centering
\caption{Baseline configuration. Every table in this paper is generated at the
baseline except in the column indicated; entries marked \emph{not reported here}
belong to experiments of the same run whose results this paper does not use.}
\label{tab:params}
\small
\begin{tabular}{llll}
\toprule
Symbol & Meaning & Baseline & Varied in \\
\midrule
$W$                  & weeks                                & $120$          & --- \\
$R$                  & regions                              & $6$            & $8$--$30$, not reported here \\
$p_0$                & reference list price                 & $5{,}200$      & --- \\
$n_{\text{moves}}$   & list-price moves per series          & $10$           & $\{2,4,6,8,12,16\}$, \S\ref{sec:frontier} \\
mag                  & move magnitude                       & $3.0$--$7.5\%$ & four ranges, \S\ref{sec:frontier} \\
$\rho_i$             & configured pass-through              & $0.85$         & $\{0,0.30,0.60,0.85\}$, \S\ref{sec:frontier} \\
$\sigma_\rho$        & regional dispersion of $\rho$        & $0$            & $0.10$--$0.18$, not reported here \\
$\lambda$            & state dependence of $\rho$           & $0$            & $0.35$--$0.45$, not reported here \\
promo\_conf          & confounding of moves with promotion  & $0$            & $\{0,0.5,0.9\}$, \S\ref{sec:frontier} \\
comp\_moves          & competitor discrete moves            & $8$            & --- \\
$\kappa$             & competitor reaction to own price     & $0.5$          & $0$ in Tables~\ref{tab:levels}, \ref{tab:hier}, \ref{tab:nivel} \\
cross                & cross-price elasticity               & $+0.30$        & --- \\
$\sigma_u$           & demand shock, logs                   & $0.05$         & --- \\
$\sigma_S$           & regional income dispersion           & $0.020$        & --- \\
$a$                  & injected asymmetry                   & $0$            & $0.25$, not reported here \\
\bottomrule
\end{tabular}
\end{table}

\paragraph{The second generator.} The cross-generator check of
Section~\ref{sec:limitations} uses the data-generating process of the baseline
data pipeline, written earlier and independently. Its structure is the same ---
\eqref{eq:pt} and \eqref{eq:demanda} with a constant pass-through of $0.85$ ---
but its topology is fixed: six units across three brands covering the five
regimes, six named regions with fixed mean temperatures, and no state-dependent
pass-through, no regional income component and no discrete competitor moves. It
is the generator behind the recovery experiment; the extended one described above
is used everywhere else.

\section{Estimators and replication counts}\label{app:reps}

\subsection{Nuisance learners and inference}

The confounder set is:
\begin{align*}
W = \{ & \text{promotional flag}, \text{promotional intensity}, \text{log competitor price}, \\
       & \text{weather}, \text{holidays in the month}, \text{working days}, \\
       & \text{week of year}, \text{log inflation} \}.
\end{align*}\textsf{gbr} is a gradient-boosted regressor (maximum depth $3$,
$200$ trees, learning rate $0.05$, subsample $0.8$) on those columns plus a week
index. \textsf{sieve} is a cross-validated ridge (penalty selected over $25$
values on a log grid from $10^{-3}$ to $10^{3}$) on a standardised explicit
basis: the confounders, a standardised week trend and its square, and three
annual harmonic pairs in the week of year. The remaining learners of
Table~\ref{tab:learners} share the sieve basis (\textsf{ridge} at a fixed
penalty, \textsf{lasso} at $\alpha = 0.001$, \textsf{rbf} a NystrÃ¶m feature map
followed by cross-validated ridge) or the tree family (\textsf{histgb},
histogram gradient boosting, depth $3$, $250$ iterations, learning rate $0.06$).

Nuisance functions are fit on $K=5$ contiguous temporal blocks with all regional
observations of a week assigned to the same block, and with no embargo between
blocks. The latter is a known limitation: a week can be split between training
and test through another region. An embargoed variant is implemented and
measured, and is not adopted here because it would change the residualiser that
two of the ablations compare.

The point estimator is the median across folds, \eqref{eq:dml}. The choice is
supported for the boosted learner and not for the sieve: over $200$ panels the
paired contrast in absolute error against the pooled aggregate is $-0.0911$
(s.e.\ $0.0152$, $t = -6.0$) for \textsf{gbr}, in favour of the median, and
$+0.0400$ (s.e.\ $0.0099$, $t = 4.0$) for \textsf{sieve}, against it. Against the
across-fold mean the median wins for \textsf{gbr} ($-0.1974$, $t = -10.2$) and
ties for \textsf{sieve} ($+0.0108$, $t = 1.0$). We report the median throughout
and record that at the aggregation levels of Section~\ref{sec:level} the
aggregate estimator is the pooled one, since a fold median is not defined across
stacked units.

Block length for the moving-block bootstrap is
$b = \mathrm{clip}(\mathrm{round}(1.5\,W^{1/3}),\,2,\,\lfloor W/3\rfloor)$,
giving $b = 7$ at $W = 120$; blocks are drawn over weeks and every regional
observation of a resampled week travels with it, except in the hierarchical
variant of Table~\ref{tab:eight}, which additionally resamples cells within each
replicated week.

\subsection{Replication counts}

Table~\ref{tab:reps} reports the experiments behind each table of the paper.
Replications are independent panels; a unit-panel is one estimated
unit $\times$ replication pair, which is the quantity that determines the
standard error of a coverage figure. The full run comprises $35$ experiments and $69.5$ hours of aggregate compute;
the remaining $24$ experiments concern components of the broader decision framework
that this paper does not report.

\begin{table}[t]
\centering
\caption{Replication counts of the experiments reported in the paper. Seconds are
aggregate compute over the replications of the experiment, executed on eight
cores. Where two scenarios are run the count is reported as replications
$\times$ scenarios.}
\label{tab:reps}
\begin{tabular}{llrrr}
\toprule
Table & Experiment & Reps & Unit-panels & Seconds \\
\midrule
\ref{tab:decomp}    & \texttt{cobertura\_condicional} & $12 \times 40$ & $480$   & $2{,}475$ \\
\ref{tab:eight}     & \texttt{inferencia\_variantes}  & $200$          & $200$   & $3{,}164$ \\
\ref{tab:frontier}  & \texttt{frontera}               & $8$            & $2{,}304$ & $10{,}573$ \\
\ref{tab:levels}    & \texttt{nivel\_identificacion}  & $200$          & $2{,}200$ & $7{,}877$ \\
\ref{tab:learners}  & \texttt{aprendiz\_matriz}       & $120$          & $120$   & $1{,}092$ \\
\ref{tab:hier}      & \texttt{jerarquia\_cobertura}   & $200 \times 2$ & $3{,}200$ & $14{,}818$ \\
\ref{tab:nivel}     & \texttt{jerarquia\_nivel}       & $100 \times 2$ & $8{,}400$ & $36{,}525$ \\
\addlinespace
App.\ \ref{app:reps} & \texttt{estimador\_puntual}    & $200$          & $200$   & $1{,}134$ \\
\ref{tab:recovery} & \texttt{recovery}           & $500$          & $500$   & $38{,}903$ \\
\S\ref{sec:limitations} & \texttt{recovery\_ext}      & $150$          & $150$   & $8{,}683$ \\
\S\ref{sec:limitations} & \texttt{calib\_vs\_eval}    & $60$           & $240$   & $1{,}172$ \\
\bottomrule
\end{tabular}
\end{table}

\paragraph{Seed separation.} Selection and reporting use disjoint seeds. The
bench that chose among interval constructions and learners ran on seeds
$20{,}000$--$24{,}000$ and the audit that verified the pre-registered rules on
$30{,}000$--$34{,}000$, while the runs reported here use bases outside both
ranges: $1{,}000$--$16{,}000$ for the experiments inherited from the pipeline
and $40{,}000$--$59{,}000$ for those introduced in this paper. A published
figure therefore does not inherit the selection bias of having chosen the best
of several variants on the same draws.

\paragraph{Pre-registered verdicts.} Every claim in this paper corresponds to a
rule written before the run, with its threshold fixed in advance, and each rule
returns pass or fail on the measured quantity. Of $30$ rules, $13$ pass, $16$
fail and one does not apply because the failure it was written to diagnose did
not reproduce. These outcomes form the empirical core of
Sections~\ref{sec:eight}--\ref{sec:level}: the claim that the baseline interval
achieves nominal coverage, the claim that some within-panel construction covers,
the claim that the shortfall is purely one of width, the claim that target coverage
is obtained at a practical width, the claim that a single learner dominates across
bias, coverage, and error, and the claim that the variance-component interval meets the precision
threshold at some level of aggregation, all fall short of their pre-registered criteria.

\clearpage
\bibliographystyle{plainnat}
\bibliography{referencias}

\end{document}